\documentclass[twocolumn]{fairmeta}

\title{Learning and Leveraging World Models in Visual Representation Learning}

\usepackage{microtype}
\usepackage{graphicx}
\usepackage{booktabs} %
\usepackage{enumitem}

\usepackage{tikz-cd}

\usepackage{hyperref}

\usepackage{amsmath}
\usepackage{amssymb}
\usepackage{mathtools}
\usepackage{amsthm}

\theoremstyle{plain}

\theoremstyle{definition}

\theoremstyle{remark}

 \usepackage{multirow}
\usepackage{colortbl}

\definecolor{Gray}{gray}{0.9}
\definecolor{Red}{rgb}{0.78, 0.1, 0.06}
\definecolor{Green}{rgb}{0.06, 0.80, 0.25}

\usepackage[textsize=tiny]{todonotes}

\author[1,2]{Quentin Garrido}
\author[1]{Mahmoud Assran}
\author[1]{Nicolas Ballas}
\author[1,3]{Adrien Bardes}
\author[2]{Laurent Najman}
\author[1,4,5]{\mbox{Yann LeCun}}

\affiliation[1]{FAIR at Meta}
\affiliation[2]{Univ Gustave Eiffel, CNRS, LIGM, F-77454 Marne-la-Vallée, France}
\affiliation[3]{INRIA}
\affiliation[4]{Courant Institute, New York University}
\affiliation[5]{Center for Data Science, New York University}

\abstract{

Joint-Embedding Predictive Architecture (JEPA) has emerged as a promising self-supervised approach that learns by leveraging a \textit{world model}. While previously limited to predicting missing parts of an input, we explore how to generalize the JEPA prediction task to a broader set of corruptions.
We  introduce Image World Models, an approach that goes beyond masked image modeling and learns to predict the effect of global photometric transformations in latent space.
We study the recipe of learning performant IWMs and show that it relies on three key aspects: conditioning, prediction difficulty, and capacity.  Additionally, we show that the predictive world model learned by IWM can be adapted through finetuning to solve diverse tasks; a fine-tuned IWM world model matches or surpasses the performance of previous self-supervised methods. 
Finally, we show that learning with an IWM allows one to control the abstraction level of the learned representations, learning invariant representations such as contrastive methods, or equivariant representations such as masked image modelling.}

\correspondence{Quentin Garrido at \email{garridoq@meta.com}}

\begin{document}

\maketitle

\section{Introduction}

Learning and leveraging world models is common practice in reinforcement learning (RL), with demonstrable success in the last few years in particular~\cite{ha2018worldmodels,hafner2019dreamer,hafner2023dreamerv3}.
World models are commonly learned by training a network to predict the consequence of an action, either in input space~\citep{yang2023unisim}, or in latent space~\citep{hu2023gaia1,hafner2023dreamerv3}.
Given such a broad view of world modelling, we seek to explore whether learning and leveraging world models can also be benificial in visual representation learning.

\begin{figure}[!t]
    \centering
    \includegraphics[width=\columnwidth]{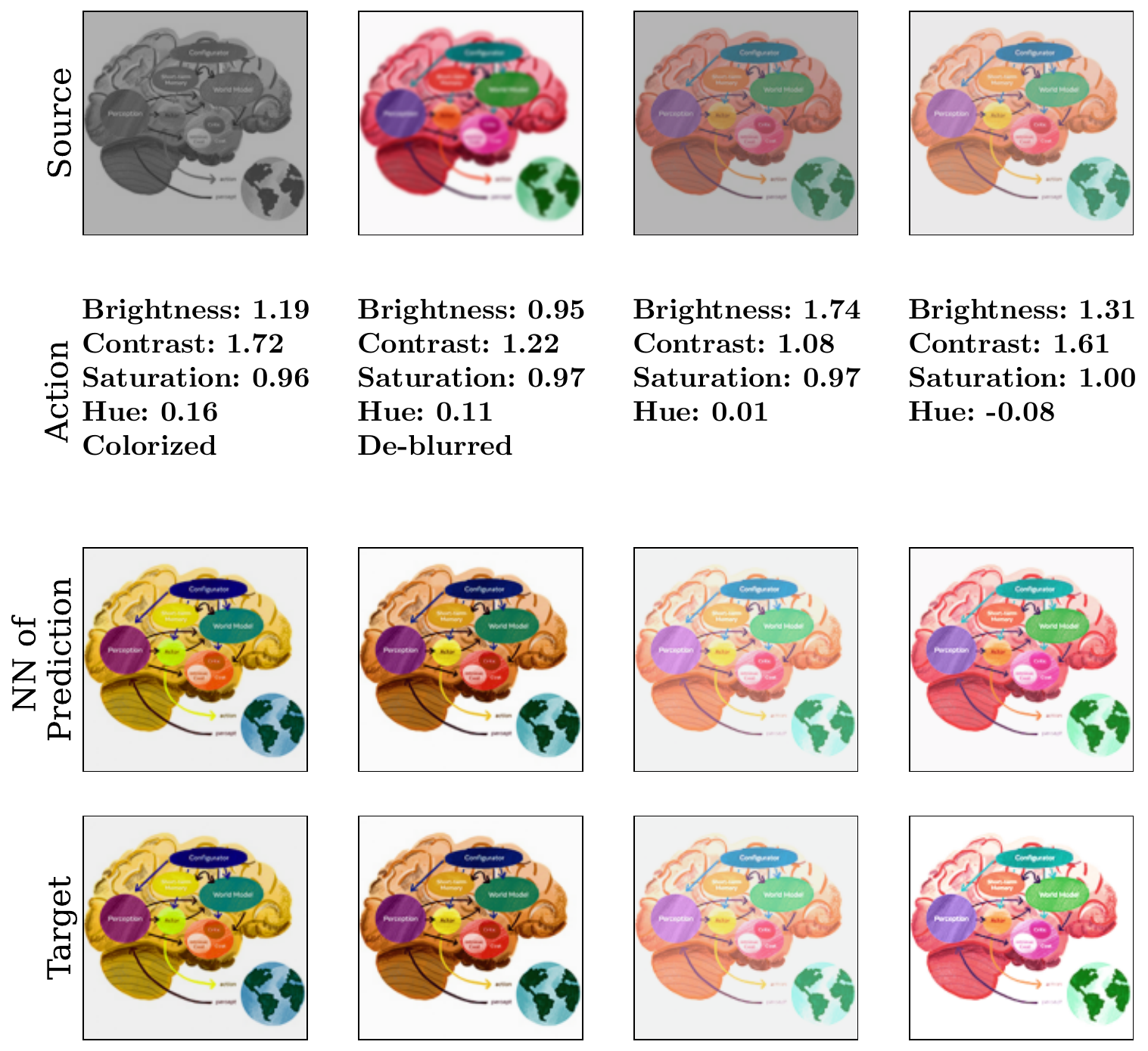}
    \vspace{-2em}
    \caption{\fontsize{8.5pt}{8.5pt}\selectfont
    \textbf{Visualisation of predictions in latent space with a learned Image World Model.} We apply an action on a source image in latent space and retrieve the nearest neighbour of the predicted representation in a bank of 256 images. We see that IWM is capable of modeling transformations and undo corruptions, showing an understanding of the underlying image transformations. Image from: \href{https://ai.meta.com/blog/yann-lecun-advances-in-ai-research/}{ai.meta.com/blog/yann-lecun-advances-in-ai-research/} 
    }
    \label{fig:iwm-visu}
    \vspace{-1em}
\end{figure}

\begin{figure*}[!t]
    \centering
    \includegraphics[width=0.9\textwidth]{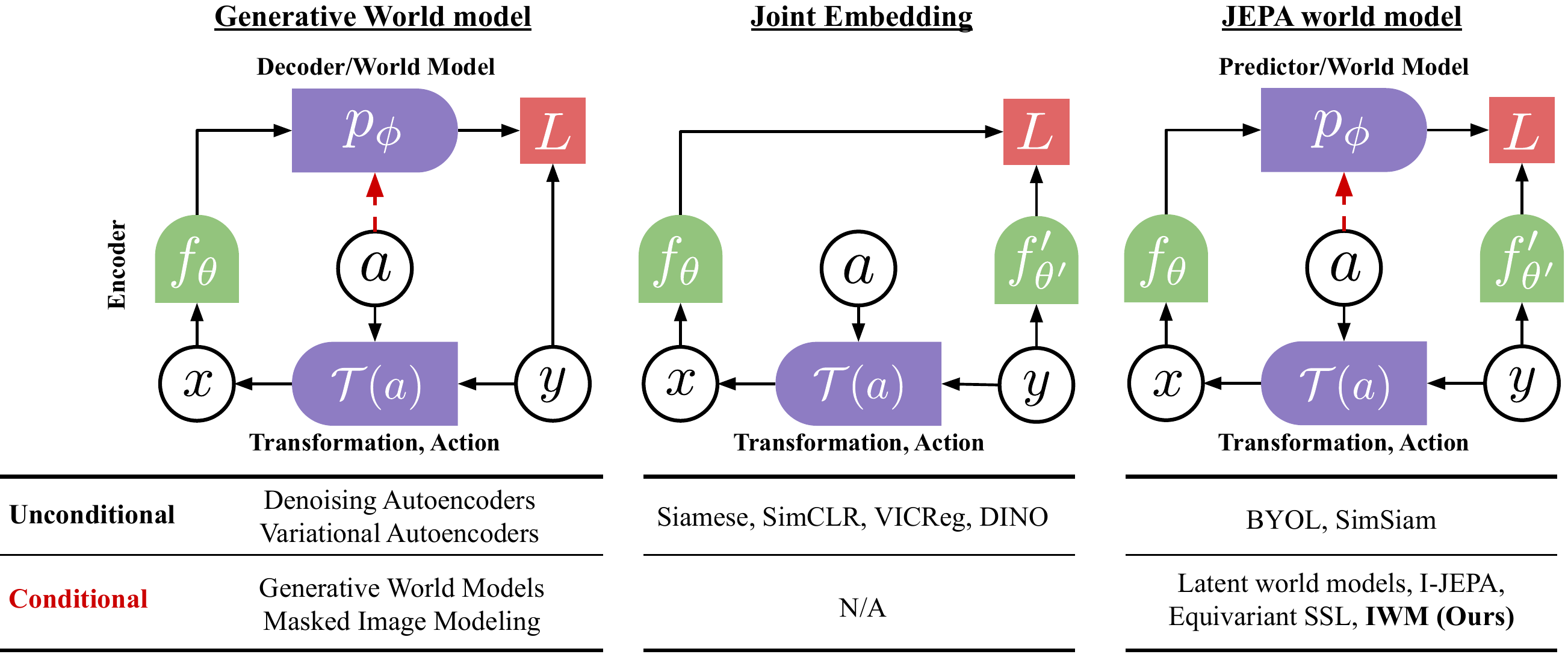} 
    \caption{\fontsize{8.5pt}{8.5pt}\selectfont
    Multiple families of methods with related architectures can be distinguished, in which the conditioning or not of their world model is a key distinction. \textbf{Generative World Models} are trained to invert a transformation in input space, leveraging an autoencoder framework. Methods for world modeling and representation learning can be instantiated in this way. \textbf{Joint Embedding} methods get rid of the world model but operate in latent space by encoding what is common between transformed inputs. It is the main class of SSL methods. \textbf{JEPA World Models} can be seen as a more general framework where a world model is trained in latent space. This family has been very successful both in reinforcement learning and in representation learning, and is where Image World Models (IWM) falls.}
    \label{fig:all-methods}
    \vspace{-1em}
\end{figure*}

A wide family of self-supervised learning approaches are based on encoder-predictor architectures, wherein the encoder-predictor networks are trained to predict transformations of the data; e.g., masked image modelling~\citep{bao2021beit,he2021mae}, joint-embedding architectures~\citep{grill2020byol, xie2022simmim, assran2023ijepa, baevski2022data2vec}, or equivariant prediction objectives~\citep{gupta2023care,garrido2023sie}.
If we regard transformations of the data as ``actions,'' then we can easily relate self-supervised learning approaches to world-modelling in reinforcement learning; see figure~\ref{fig:all-methods}.

For instance, the decoder network in masked autoencoders~\citep{he2021mae} can be thought of as a generative image world model, which learns to infer the effect of the ``masking action'' $\mathcal{T}(a)$ on an image $y$; in this case, the transformation parameters $a$ (locations of masked image patches), are also fed to the decoder network.
Methods based on joint-embedding predictive architectures (JEPAs), such as I-JEPA~\citep{assran2023ijepa} or data2vec~\citep{baevski2022data2vec}, operate similarly, but can be seen as learning a latent image world model, which learns to infer the effect of the masking action on the representation of an image.
If one does not condition the predictor on the transformation parameters, then the best we can hope for is learning representations that are invariant to the data transformations, as in BYOL~\citep{grill2020byol} and SimSiam~\citep{chen2020simsiam}, wherein the image transformations correspond to various photometric and geometric data augmentations.\\
However, despite some of the apparent similarities between world modelling in reinforcement learning and self-supervised learning from images, the learned world model in reinforcement learning is typically leveraged in downstream tasks, e.g., for planning~\citep{hansen2022modem}.
In contrast, the learned world model in self-supervised learning is typically discarded after pretraining, as the main focus is often on the representation quality of the learned encoder network.
This stems from the fact that most downstream tasks in computer vision are unrelated to the world modeling task.
Common tasks of interest focus on discriminative aspects and as such, even when the predictor learns useful information, it is simply discarded.
We postulate that discarding the world model in representation learning is wasteful, and that just like in RL, we can reuse this world model for downstream tasks.
This motivates us to study, in more depth, learning world models as a paradigm for representation learning. 
We thus introduce Image World Models (IWM, illustrated to the right of figure~\ref{fig:all-methods}) as a way to learn both good representations and strong reusable world models.
IWM is based on JEPA and extends the usual latent inpainting to also include photometric transformations, allowing us to demonstrate the key aspects in learning a capable world model, which include the choice of predictor conditioning, the strength of the transformations, and the capacity of the world model.\\
We then focus on leveraging the learned world model for downstream tasks, and find that it can be leveraged through finetuning.
Specifically, we find that finetuning the world model on top of the frozen encoder for downstream tasks provides improved performance over encoder finetuning; this is also achieved at a fraction of the cost and number of finetuned parameters.
Moreover, only the world model learned by IWM exhibits this behavior; finetuning a randomly initialized network of the same architecture as the predictor does not provide such a performance improvement.
This suggests that the world model should be a key part of the inference process, instead of being discarded.
Inspired by instruction tuning~\citep{wei2022instruction,zhang2023instruction}, we further show that the world model can be finetuned to solve multiple tasks at once, further improving efficiency.

Our study reveals another key aspect of representation learning with world models: the capacity given to the world model has a direct influence on the level of abstraction of the learned representations.
Intuitively, if the predictor is the identity (i.e., no predictor, middle of figure~\ref{fig:all-methods}), the network will capture high level semantic information, as it will only learn to encode what is shared between the input $y$ and its transformation $x$.
This is the driving force behind the representation quality of contrastive learning, where transformations are selected to only preserve the semantics of the image.
On the other hand, as the predictor has more capacity and can effectively invert the effect of the transformations, the output of the encoder can retain more information about its input.
These two ideas are at the core of equivariant representation learning; a predictor that can apply transformations effectively is equivariant, whereas a predictor that cannot is invariant.
We find that a world model that is invariant to transformations performs better in linear evaluation, whereas one that is equivariant correlates with better world model finetuning.
This gives a tradeoff between ease of adaption and raw performance.
As such, learning representations by learning a world model gives us flexibility in the properties of the representations, making this an attractive representation learning framework.\\
Our contributions can be summarized as follows:
\begin{itemize}[noitemsep,nosep]
    \item We show how to leverage JEPAs to learn an Image World Model (IWM). The key aspects are: complexity of transformations, conditioning on transformations, and capacity of the predictor.
    \item We show that equivariant world models can be leveraged for discriminative tasks. Finetuning the predictor leads to better performance compared to encoder finetuning, at a fraction of the cost. Inspired by instruction tuning, we also demonstrate that it can be finetuned on several tasks at once.
    \item We show that controlling the capabilities of the world model gives us representations with different properties. An invariant world model gives us more abstract representations and performs better in linear evaluation, akin to contrastive learning. An equivariant world model preserves more information about the input, giving better peak performance with predictor finetuning.
\end{itemize}

\section{Related works}

\subsection{Augmentation invariant Self-Supervised Learning}
At the core of contrastive methods lies augmentation invariance. Multiple augmented views of an image should lead to the same representation in latent space. The core of these methods is thus in how to avoid these representations collapsing. Sample-contrastive methods~\citep{chen2020simple,he2020moco,chen2020mocov2,caron2021dino,chen2021mocov3,yeh2021decoupled,haochen2021provable,oquab2023dinov2} avoid this phenomenon by pushing away representations coming from other data points. Dimension-contrastive methods~\citep{bardes2021vicreg,zbontar2021barlow,ermolov2021whitening,li2022neural,bardes2022vicregl} avoid collapse by considering the representations as a whole and encouraging maximization of information content. Both dimension- and sample-contrastive methods have been shown to lead to very similar representations~\citep{garrido2023duality}.
Prediction based methods~\citep{grill2020byol,chen2020simsiam} learn by predicting the augmented representations, but they also lead to invariant representations due to a lack of conditioning on the transformations.

\subsection{World modeling in visual representation learning}
While world modeling is a successful paradigm in reinforcement learning~\cite{hafner2019dreamer,hafner2023dreamerv3} or video prediction~\cite{yang2023unisim,hu2023gaia1}, it has yet to show clear benefits in representation learning.
However, multiple families of approaches can be reframed in light of this.
Equivariant self-supervised learning methods~\citep{devillers2022equimod,park_learning_2022,garrido2023sie,gupta2023care,dangovski_equivariant_2022} aim to predict transformations of data when such transformations form a group.
Masked Image Modeling~\cite{he2021mae,bao2021beit,el2024aim,xie2022simmim} learns representations by predicting masked parts of the image. While these approaches predict in pixel space, their decoders can be seen as instantiations of world models. Similarly, JEPAs~\citep{assran2023ijepa,baevski2022data2vec} predict masked parts of the image, but in the latent space.
Recently, generative approaches have been applied to representation learning~\cite{hudson2023soda,clark2023text,chen2024dae}, and while these approaches seem promising, their performance still remains below contrastive or MIM approaches. Recent work has also shown negative correlations between generation quality and representation quality~\citep{chen2024dae}.
One shared aspect among these works is that the world model (predictor or decoder) is either discarded for evaluations, or only used to augment data~\citep{hudson2023soda}. We propose to go beyond these practices and show that we can learn a world model that is reusable for downstream tasks while still learning high-quality representations.

\section{Method}

We now describe Image World Models (IWM). It follows a Joint Embedding Predictive Architecture framework~\citep{lecun2022AMI} akin to I-JEPA~\citep{assran2023ijepa}. In this framework, the predictor is the instantiation of the world model. We consider a world model to be capable if it can apply transformations in latent space, and thus learns equivariant representations. As such, we call a capable world model equivariant~\footnote{This is an abuse of language as not all considered transformations form a group, but it is used for clarity.} and a poor world model invariant.\\
An appealing aspect of using JEPAs is that approaches which learn equivariant representations using contrastive methods often have to rely on an invariance loss to increase representation quality, whether explicitly~\citep{gupta2023care,garrido2023sie}, or implicitly~\citep{chavhan2023diversity}. On the other hand, a JEPA style approach does not have this drawback, as the semantic aspect of the representation is learned through latent inpainting.
Working in latent space further allows the network to remove unnecessary information, or that which is too hard to predict. This makes the JEPA formulation attractive since, for reconstructive methods, the quality of the reconstruction is not necessarily correlated with representation quality~\cite{chen2024dae}. \\
To train IWM, the first step is to generate source and target views --- $x$ and $y$ respectively in figure~\ref{fig:all-methods} --- from an image $I$.\\

\textbf{Target $y$.} The target view is generated by applying a random horizontal flip, a crop, and color jitter (brightness, contrast, saturation, hue) to the original image $I$. No destructive augmentations such as grayscale are applied on the target to ensure that the target has as much information as possible. We further elaborate on this choice in appendix~\ref{app:data-aug}.

\textbf{Source $x$.} For the source view, we start from the target $y$ which we further transform. We first apply another color jitter, as well as destructive augmentations: grayscale, blur and solarization. This set of augmentations is the same as the one used in contrastive SSL. Finally, we also mask parts of the image following I-JEPA. We define our mask $M_x$ (a set of indices) as the union of 4 rectangular masks. Confer appendix~\ref{app:exp-details} for exact implementation details.

\textbf{Action $a$.} We denote by $a_{x\rightarrow y}$ the transformation parameters associated with the transformation of $x$ to $y$, i.e., the invert of the initial transformation process. $a_{x\rightarrow y}$ contains information about the color jitter difference between $x$ and $y$ as well as information on whether or not each destructive augmentation was applied.

\textbf{World modeling with $p_\phi$.}
The source and target are then fed respectively through an encoder $f_\theta$ and its exponential moving average $f_\theta^\text{EMA}$. This gives us representations $z_x = f_\theta(x)$ and $z_y = f_\theta^{\textbf{EMA}}(y)$. The use of the EMA network is crucial to avoid collapsed solutions.
To condition the predictor, acting as our world model, it is fed with geometric information about the target in the form of mask tokens as well as $a_{x\rightarrow y}$. We denote these mask tokens as $m_a$, which correspond to the positions in $M_x^C$.
The predictor $p_\phi$ then takes as input the embedded source patches $x_c$, transformation parameters $a_{x\rightarrow y}$ and mask tokens $m_a$. Its objective is then to match $p_\phi\left(z_x,a_{x\rightarrow y},m_a \right) = \hat{z_y}$ to $z_y$.

\textbf{Loss.} The loss function used is a squared $L2$ distance between the predictions $\hat{z_y}$ and their targets $z_y$: 
\begin{equation*}
    L(x,y) = \sum_{i\in M_x^C}\| p_\phi\left(f_\theta(x),a_{x\rightarrow y},m_a \right)_i
    - f_\theta^\text{EMA}(y)_i \|_2^2.
\end{equation*}

\subsection{Architecture and nomenclature}
Our encoder is a Vision Transformer~\citep{dosovitskiy2021vit}, in particular we use the ViT-B/16 architecture. Our predictor is based on the same architecture with different depth and embedding dimension. We denote instances of IWM as $\text{IWM}_{X,Y}^Z$ where $X$ is the depth of the predictor, $Y$ its embedding dimension, and $Z$ is either Inv or Equi depending on the capabilities of the world model. For example $\text{IWM}_{18,384}^\text{Equi}$ means that the predictor is 18 layers deep, with 384 dimensional embeddings and exhibits equivariant behavior, i.e., has learned a versatile world model.

\section{Learning an Image World Model for representation learning}

\subsection{Evaluating the quality of the world model}

As discussed previously, learning equivariant representations and learning a world model are closely related problems. As such, we can borrow metrics from the equivariance literature to evaluate the quality of a trained world model. We rely on Mean Reciprocal Rank (MRR)~\citep{kipf2019contrastive} as our main metric. To compute it, we generate a bank of augmented target images (256 in practice). We feed the representation of the clean image through the predictor with the goal of predicting the target image. We then compute the distance between the prediction and the augmented representation bank from which we get the rank of the target in this NN-graph. Averaging the reciprocal ranks over multiple images and transformations gives us MRR which tells us about the quality of the world model. A MRR close to 1 means that the world model is able to apply the transformation, on the contrary a MRR close to 0 means that it cannot. 

\subsection{Learning a strong Image World Model}

In order to build a performant IWM, we isolate three key aspects: conditioning the predictor on transformations (or actions), controlling the complexity of the transformations, and controlling the capacity of the predictor. We show that not caring properly for either of those leads to invariant representations.

\textbf{World model conditioning.} We study two approaches to condition the predictor on the transformation information. \\
\textit{Sequence conditioning.} One approach is simply to add tokens representing the transformation to the input of the predictor. Although this seems straightforward, it needs to be implemented in a way that breaks the permutation equivariance of the transformer predictor. To do so, every token is fed through a unique linear layer that allows the network to transform the information in a way that can be disambiguated by the predictor. \\
\textit{Feature conditioning.} Another option is to mix the information between the transformation and mask tokens by adding the conditioning as extra dimensions, then feeding the mask tokens through a 1x1 convolutional neural network to mix the information in the mask tokens and map back to the right dimensionality. \\
As we can see in Table~\ref{tab:conditioning}, no conditioning leads to a world model that cannot apply transformations whereas both conditioning using the sequence or feature axes leads to good world models. We use the feature conditioning in practice as it leads to higher downstream performance.

\begin{table}[!t]
    \centering
    \caption{\fontsize{8.5pt}{8.5pt}\selectfont
    \textbf{Influence of predictor conditioning on the quality of the world model.} Both Sequence and Feature conditioning lead to good world models .\textcolor{gray}{Gray} is our default setting.}
    \begin{tabular}{lccc}
    \toprule
      \it Conditioning: & None & Sequence & Feature \\
     \midrule
     MRR & 0.00 & 0.82 & \cellcolor{Gray} 0.79 \\
     \bottomrule
    \end{tabular}
    \label{tab:conditioning}
    \vspace{-1em}
\end{table}

\begin{table}[!t]
    \centering
    \caption{\fontsize{8.5pt}{8.5pt}\selectfont
    \textbf{Impact of predictor architecture and transformations on MRR.} Learning an effective world model requires complex transformations and adequate predictor capacity. \textcolor{gray}{Gray} is our default setting. \textcolor{Red}{Red} and \textcolor{Green}{Green} respectively indicate invariant and equivariant behavior.}

    \begin{tabular}{lccc}
    \toprule
     \it Predictor: & I-JEPA     & \multicolumn{2}{c}{IWM} \\
     \cmidrule(l){2-2}  \cmidrule(l){3-4}
     \it (depth, dim.): & (12,384) & (12,384) & (18,384)  \\
     \midrule
    Jitter & \textcolor{Red}{0.00} &\textcolor{Red}{0.11} & \textcolor{orange}{0.25} \\
    + Destructive  & \textcolor{Red}{0.00}   & \textcolor{Red}{0.09} & \cellcolor{Gray} \textcolor{Green}{0.79}  \\
    + Strong Jitter & \textcolor{Red}{0.00} & \textcolor{Green}{0.81} & \textcolor{Green}{0.85} \\
     \bottomrule
    \end{tabular}
    \label{tab:achieving-equivariance}
\end{table}

\textbf{Transformation complexity.} We rely on data augmentation as used in contrastive approaches, consisting of color jitter (brightness, hue, contrast, saturation), grayscale, blur, and solarization. We refer to the last three as destructive since they remove information.
Beyond the set of transformations modeled, their strength must also be adequate to learn a useful world model. If the prediction task is too easy, then the predictor will not learn anything useful. As presented in Table~\ref{tab:achieving-equivariance}, the stronger the augmentations, the easier it is to learn a strong world model. We provide more detailed ablations on the augmentations in Appendix~\ref{app:data-aug}, where we see the trend continuing on a wider range of augmentation scenarios.

\textbf{World model capacity.} If the transformation is complex, the predictor needs more capacity to be able to apply it, motivating capacity as a crucial factor in learning Image World Models.
As we can see in Table~\ref{tab:achieving-equivariance}, a deeper predictor enables us to learn a strong world model on a wider range of augmentations, and is key to the success of IWM. We study in more detail the influence of depth on achieving a good world model in appendix~\ref{app:data-aug}. For 12 layers, jitter equivariance is achieved 1 out of 5 times whereas for the 18 layers, it is achieved 4 out of 5 times. As such, predictor capacity is a key component of a strong world model.%

\subsection{Visualizing predictions.} In the same way that we computed MRR, we can compare the predicted representations to a bank of transformed images and look at the image associated to the prediction's nearest neighbor. As we see in Figure~\ref{fig:iwm-visu} the world model learned by IWM is able to properly apply transformations in latent space. We can however see some inaccuracies when inverting grayscale as it is not properly invertible.
These visualisations help reinforce the fact that IWM is able to learn strong world models for image transformations. Confer appendix~\ref{app:more-qualitative} for more visualizations.

\section{Leveraging world models for downstream tasks \label{sec:pred-ft}}

A limitation of world models learned on images is that the task they solve is not aligned with most downstream tasks. We showed that IWM can apply color jitter or colorize images, but these are not the tasks that drive applications of computer vision. This is in contrast with LLMs where predicting the next token is one of the main applications of such models. We thus study how to leverage a world model in vision, for tasks that go beyond applying transformations. We focus on discriminative tasks such as image classification and image segmentation.

\subsection{Predictor finetuning}

\begin{table}[!t]
\centering
    \caption{\fontsize{8.5pt}{8.5pt}\selectfont
    \textbf{How to predict for predictor finetuning}. Using the teacher improves performance, and the exact prediction task is not crucial. Null latents are more flexible and perform better. For better efficiency, a full prediction is not needed but leads to a small drop in performance. \textcolor{gray}{Gray} is our default setting.}
    \label{tab:pred-ft-ablation}
\resizebox{\columnwidth}{!}{\begin{tabular}{lcc}
\toprule
Setting & ImageNet Top-1 (\%) & Gap \\
\midrule
Default & 82.9 &  - \\
+ Teacher & 83.2 & + 0.3 \\
\rowcolor{Gray}  
+ Null latents & 83.3 & + 0.1 \\
\midrule
+ Pred only one token & 82.8 & -0.5 \\
\bottomrule
\end{tabular}}
\vspace{-1em}
\end{table}

\begin{table*}[!t]
\centering
    \caption{\fontsize{8.5pt}{8.5pt}\selectfont
    \textbf{Finetuning evaluations on ImageNet-1k.} We evaluate prediction based methods by finetuning their encoder, by keeping the encoder frozen and finetuning their predictive world model or by finetuning both. Finetuning the world model is highly effective with IWM when it exhibits an equivariant behavior. This behavior is absent or less clear with other methods, showing the importance of a strong world model. }
    \label{tab:full-results}
\begin{tabular}{lccccc}
\toprule
\multirow{2}{*}{Method} & \multirow{2}{*}{Epochs} & \multicolumn{1}{c}{No predictor} & \multicolumn{2}{c}{Frozen encoder, tuned predictor}  & \multirow{2}{*}{End to end}  \\
\cmidrule(l){3-3} \cmidrule(l){4-5}
& & Encoder & Random Init. & Pretrained &   \\
\midrule
\multirow{ 2}{*}{MAE} & 300  & 82.7 & 82.4 & 82.7 \textcolor{Red}{(+0.3)} & 82.3  \\
& 1600  & \textbf{83.6} & \textbf{83.0} & 83.1 \textcolor{Red}{(+0.1)} & 83.3  \\
I-JEPA & 300 & 83.0 & 79.1 & 80.0 \textcolor{orange}{(+0.9)} & 82.0   \\
\midrule
$\text{IWM}_{12,384}^\text{Inv}$ & 300  & 83.3 & 80.5 & 81.3 \textcolor{orange}{(+0.8)} & 82.7  \\
$\text{IWM}_{18,384}^\text{Equi}$ & 300 & 82.9 & 81.5 & \textbf{83.3} \textcolor{Green}{(+1.8)} &  \textbf{84.4} \\

\bottomrule
\end{tabular}
\vspace{-1em}

\end{table*}

For any task, the evaluation head needs to understand the learned latent space and leverage it to solve the problem at hand. This is something our learned predictor can do, suggesting that it has learned useful information that is not necessarily present in the encoder. 
However, since the predictor is trained to predict another valid representation, its output has no reason to lead to better downstream performance if used as is.
This is why the predictor needs to be finetuned to solve discriminative tasks.
We thus focus on comparisons with finetuning protocols, following~\cite{he2021mae}. All methods studied are pretrained and evaluated on ImageNet~\cite{deng2009imagenet} and use ViT-B/16 as encoders.

\textbf{Prediction task.} When finetuning the predictor, we still need to use it for a prediction task. In Table~\ref{tab:pred-ft-ablation}, we study various ways to define the prediction task and how it impacts performance. The first aspect we notice is that using the teacher network improves performance over the student. Using a random transformation or not is not an important factor, and the most important one is to predict another full image. This makes the evaluation more flexible as we do not have to reuse the pretraining objective for our evaluation.
Using a CLS token to aggregate information instead of a full image prediction is also a valid strategy, though it lowers the performance by half a point. This techniques has the advantage of being cheaper ($N+1$ tokens vs $2N$) so it can be a good alternative depending on the use case. Overall, the simplest approach is the best: predicting an untransformed version of the full image. This makes the finetuning protocol easily reusable as it is not dependent on the pretraining task. 
We provide more detailed ablations in appendix~\ref{app:pred-task}.

\begin{figure}[!t]
    \centering
    \includegraphics[width=0.85\columnwidth]{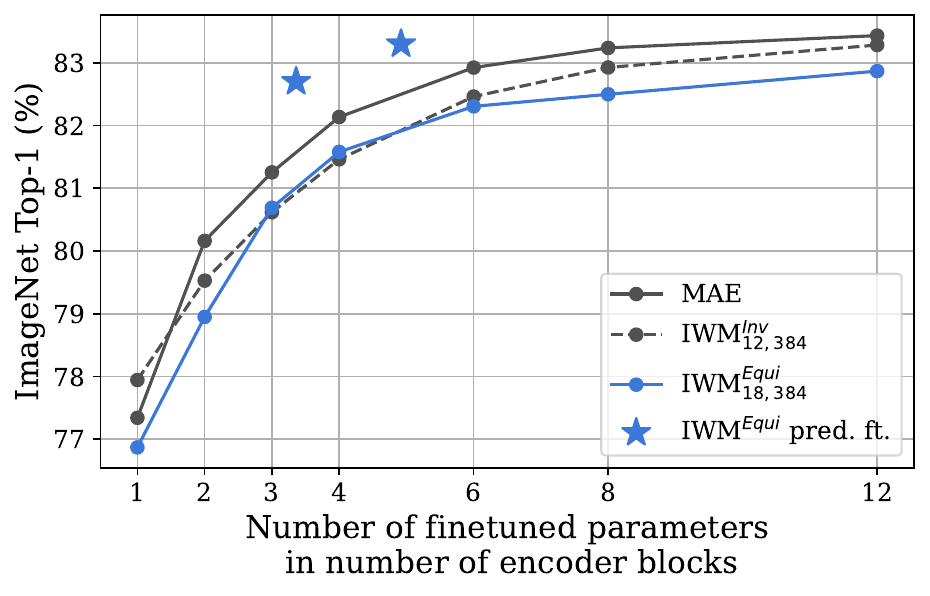}
    \vspace{-1em}
    \caption{\fontsize{8.5pt}{8.5pt}\selectfont
    \textbf{Finetuning efficiency.} When taking into account the number of finetuned parameters, predictor finetuning is significantly more efficient than finetuning the encoder.}
    \label{fig:efficiency}
    \vspace{-1em}
\end{figure}

\textbf{General Results.} In Table~\ref{tab:full-results}, we compare predictor finetuning to encoder finetuning and end-to-end finetuning of both the predictor and encoder, using ViT-B/16 for the encoder. We see that IWM maintains or improves performance over I-JEPA and that an invariant behavior is better in encoder finetuning. Interestingly, predictor finetuning of the equivariant IWM is able to match the performance of finetuning of the invariant model's encoder. This shows that the protocol can be competitive as it trades parameters at inference time for a more computationally friendly adaptation. While this evaluation increases the number of parameters used at inference time, it still amortizes the forward pass through the backbone, something that full finetuning does not do. As such, as soon as multiple tasks are considered, using the finetuned predictor provides a higher throughput than regular finetuning. \\
When comparing the use of a randomly initialized predictor (i.e., a large evaluation head) versus a pretrained predictor, we see negligible gains for MAE. This suggests that the world model learned by MAE is not better than a randomly initialized network for classification. For I-JEPA and IWM with an invariant world model, we see gains in performance lower than 1 point, suggesting that the world model is not powerful enough to be leveraged. However, when looking at IWM with an equivariant world model, we see a gain of 1.8 points over a random predictor. This shows that the predictor has learned useful information and properties that bring additional benefit to what the encoder has learned. \\
The performance can be pushed further by finetuning end-to-end both the encoder and predictor, and IWM is able to outperform every other finetuning protocols. This allows us to get more performance out of a single pretraining since the world model is always trained.
We hypothesize that the lack of performance for most approaches on end-to-end finetuning comes from the optimization complexity of finetuning a part of the network (encoder) while training from scratch another part (the predictor).
We see in Table~\ref{tab:results_peak} that when aggregating the performance over all protocols, leveraging our IWM leads to the best performance with a frozen encoder, that is when allowed to leverage every part of the pretraining. Confer Appendix~\ref{app:exp-details} for detailed performances.

\begin{table}[!t]
\centering
    \caption{\fontsize{8.5pt}{8.5pt}\selectfont
    \textbf{Peak performance achieved from a single pretraining instance.} We compare ImageNet Top-1 accuracy with a frozen encoder or when allowing any evaluation head with any protocol, finetuning or not, with a predictor on top of the encoder or not.}
    \label{tab:results_peak}
\resizebox{\columnwidth}{!}{\begin{tabular}{lccc}
\toprule
Method & Epochs & Frozen Encoder & Any protocol  \\
\midrule
DINO & 1600 & 82.0 & 82.8 \\
MOCOv3 & 300 & 76.4 & 83.2  \\
iBOT& 1600 & 83.0 & 84.0  \\
MAE & 1600  & 83.1 & 83.6 \\
I-JEPA & 300 & 80.0 & 82.0 \\
 \midrule
$\text{IWM}_{12,384}^\text{Inv}$ & 300  & 81.3 & 83.3 \\
$\text{IWM}_{18,384}^\text{Equi}$ & 300 & \textbf{83.3} & \textbf{84.4} \\

\bottomrule
\end{tabular}}
\vspace{-1em}
\end{table}

\textbf{Image Segmentation.} We study in Table~\ref{fig:results-seg} the performance of I-JEPA and IWM on an image segmentation task on ADE20k. We observe similar trends as in image classification where the invariant model leads to the best encoder. However, finetuning the predictor with an equivariant model leads to significant gain over it, outperforming encoder finetuning by a large margin. Again, we observe gains in end-to-end finetuning. This further validates the potential of our IWM to be leveraged for a wide range of tasks. We provide additional details in Appendix~\ref{app:seg}.

\begin{table}[!t]
\centering
    \caption{\fontsize{8.5pt}{8.5pt}\selectfont
    \textbf{Finetuning for segmentation on ADE20k}. Similar to image classification, we observe that predictor finetuning improves performance and outperforms encoder finetuning.}
    \label{fig:results-seg}
\begin{tabular}{lccc}
\toprule
Method  & Encoder & Predictor & End to end \\
\midrule
I-JEPA &  44.2 & 45.4 & 45.1  \\
$\text{IWM}_{12,384}^\text{Inv}$  & \textbf{45.6} & 45.7  & 46.5 \\
$\text{IWM}_{18,384}^\text{Equi}$  &  44.2 & \textbf{46.8}  & \textbf{47.0}  \\

\bottomrule
\end{tabular}

\end{table}

\textbf{Efficiency.} In Figure~\ref{fig:efficiency}, we study the efficiency of predictor finetuning compared to encoder finetuning. We see that when the number of parameters is comparable, and at multiple predictor sizes, predictor finetuning with IWM outperforms encoder finetuning by around 1 point compared to MAE, and by 1.5 points over IWM. This means that predictor finetuning is not only is a competitive protocol performance wise, but also with respect to efficiency of adaptation. \\
We further study the behavior of IWM with a ViT-L/16 in section~\ref{app:large}. When comparing the end-to-end finetuning of a ViT-B with encoder finetuning of a ViT-L, we observe a gain in performance (84.4\% vs 84.3\%) with a fraction of the parameters (121M vs 307 M). This further shows how efficient leveraging the world model learned by IWM is, and that reusing all parts of your pretraining can prove as effective as scaling the encoder.

\subsection{Multitask predictor tuning}

We previously discussed efficiency gains when compared to encoder finetuning, but can improve efficiency even further. One of the main goal of representation learning is to obtain representations that can be used for a variety of tasks. And just like the predictor was trained to solve a variety of task (colorization, inpainting, changing color) we show that it can be finetuned on multiple tasks, inspired by prefix tuning~\citep{li2021prefixtuning} and instruction tuning~\cite{wei2022instruction,zhang2023instruction} in LLMs.\\
The general idea, that we illustrate graphically in supplementary Figure~\ref{fig:multitask}, is to give new learned tokens to the predictor to indicate which task it is trying to solve. This is reminiscent of DyTox~\cite{douillard2021dytox} which uses task tokens for continual learning. For each task, we thus have a task token, as well as a task specific head and/or loss function. All of the task losses are then combined, and the predictor, as well as task specific heads, are updated. We study a simple scenario where the batch is evenly split between tasks, noting that other sampling strategies may lead to further improved performance.

\begin{table}[!t]
\centering
    \caption{\textbf{Multi-task finetuning.} Finetuning the predictor on multiple tasks at once performs similarly as finetuning it on each task separately. This enables the use of a single prediction head for multiple task, amortizing its cost.}
    \label{tab:multitask}
\begin{tabular}{lccc}
\toprule
Dataset & Single-task & Multi-task & Difference \\
\midrule
ImageNet & 80.8 & 79.6  & \textcolor{Red}{-1.2}  \\
iNat18 & 72.4 & 72.0 & \textcolor{orange}{-0.4} \\
SUN397 & 75.6 & 78.2 & \textcolor{Green}{+2.6}  \\
Places205 & 64.8 & 64.1  & \textcolor{orange}{-0.7}  \\
\midrule
Average & 73.4 & 73.5 & \textcolor{orange}{+0.1} \\
\bottomrule
\end{tabular}
\end{table}

We evaluate in Table~\ref{tab:multitask} $\text{IWM}_{18,384}^\text{Equi}$ (pretrained on ImageNet) on ImageNet, iNaturalist18~\citep{vanhorni2018naturalist}, SUN397~\citep{xiao2010sun}, and Places205~\citep{zhou2014places}. For each task we train a single-task baseline where the total number of iterations is identical to the multi-task training. As such, training all four single-task baselines has exactly the same cost as the multi-task, although it leads to four different models instead of one. The multi-task predictor is able to achieve similar performance as the single-task predictors, with a moderate drop on most tasks but a significant increase in performance on SUN397. On average it achieves the same performance as the single-task predictors. This further demonstrates the efficiency gains of leveraging good world models, where the parameters are now shared across all tasks, making predictor finetuning lightweight at inference time for every task.

Overall, when a good world model is learned, it can be reused for downstream tasks by finetuning it. This leads to performance rivaling with encoder-finetuning at a fraction of the cost. It can be made even more efficient by doing a multi-task finetuning, highlighting the versatility of this approach.

\section{Image World Models enable flexible representations}

To complete our analysis of IWM for representation learning, we study how it performs on lightweight evaluation protocols that are commonly used in self-supervised learning. We focus on linear~\cite{chen2021mocov3} and attentive probing~\cite{chen2023context}.

\begin{table}[!t]
\centering
    \caption{\fontsize{8.5pt}{8.5pt}\selectfont
    \textbf{Linear and attentive probing performance on ImageNet-1k.} $\text{IWM}^\text{Inv}$ performs similarly to contrastive methods and  $\text{IWM}^\text{Equi}$ to mask modeling ones.}
    \label{tab:lin-att}
\begin{tabular}{lccc}
\toprule
Method & Effective Epochs & Linear & Attentive \\
\midrule
MoCoV3 & 300  & \textbf{76.3} & 76.4  \\
MAE & 300  & 60.2 & 73.5 \\
MAE & 1600  & 68.0 & 76.0 \\
I-JEPA & 300 & 70.0 & 75.0 \\
\midrule
$\text{IWM}_{12,384}^\text{Inv}$ & 300  & 74.5 & \textbf{77.0} \\
$\text{IWM}_{18,384}^\text{Equi}$ & 300 & 67.5 & 75.1 \\
\bottomrule
\end{tabular}
\vspace{-1em}
\end{table}

\begin{figure}[!t]
    \centering
    \includegraphics[width=1\columnwidth]{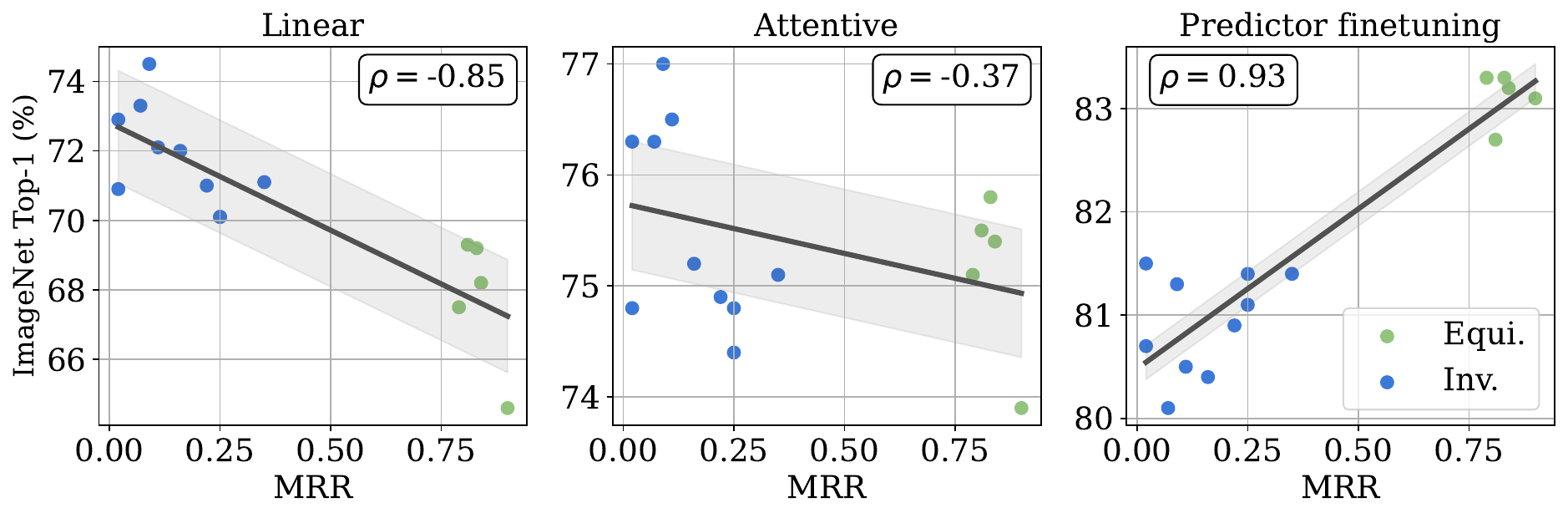}
    \vspace{-2em}
    \caption{\fontsize{8.5pt}{8.5pt}\selectfont
    While the level of equivariance influences performance in Linear and Predictor finetuning setting, it is hardly correlated to Attentive probing. This suggests that there is a trade-off in terms of the level of abstraction of the representation, and that different evaluation protocols evaluate different properties.}
    \label{fig:equi-inv-tradeoff}
\end{figure}

\begin{figure}[tb]
    \centering
    \includegraphics[width=\columnwidth]{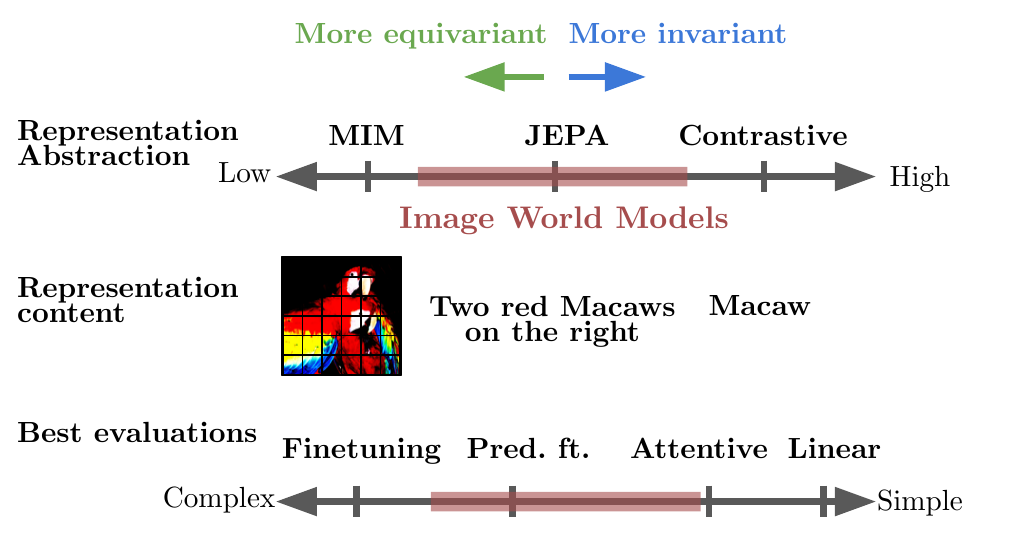}
    \vspace{-2em}
    \caption{\fontsize{8.5pt}{8.5pt}\selectfont
    \textbf{Image World Models allow representation modularity}. Different families of methods offer representations with different properties, but IWM allows exploring the whole spectrum.}
    \label{fig:abstraction}
    \vspace{-1em}
\end{figure}

As we see in Table~\ref{tab:lin-att}, when IWM learns an invariant world model, it achieves a behavior akin to contrastive approaches such as MoCov3 with significant performance gains in linear evaluation compared to MIM or other JEPA based approaches. Similarly, when IWM learns an equivariant world model, its behavior is akin to MIM methods such as MAE with lower performance in linear evaluation but more competitive performance in attentive probing.\\
This suggests that a big difference between methods is not necessarily in the quality of the representation but in their abstraction level, i.e., how easy it is to extract information from them. Linear probing being one of the simplest evaluations, attentive being slightly more elaborate and finetuning being a more complex protocol.\\
In Figure~\ref{fig:equi-inv-tradeoff}, we see clear links between most suited evaluation protocols and equivariance of the world model. More invariant world models excel in linear evaluation and equivariant world models shine with larger evaluation heads such as in predictor finetuning. We also note that the richer representations stemming from equivariant world models lead to better performance on OOD datasets(see appendix~\ref{app:ood}).\\
This allows us to place families of approaches on a spectrum of representation abstraction in Figure~\ref{fig:abstraction}. Contrastive methods occupy the high abstraction end of the spectrum, with information that is easily extractible with a simple protocol. However they suffer from lower peak performance when ignoring the adaptation cost, as seen in Table~\ref{tab:results_peak}.
On the opposite end lies Masked Image Modeling, which offers stronger performance with complex evaluations such as finetuning but suffers in linear probing as information is not as easily accessible. By varying the equivariance of the world model, IWM is able to occupy the spectrum in between contrastive approaches and MIM, as we can see in Figure~\ref{fig:equi-inv-tradeoff} and Table~\ref{tab:lin-att} with IWM$^\text{Inv}_{12,384}$ and IWM$^\text{Equi}_{18,384}$ being the two extremes of the IWM spectrum.\\
This spectrum can be summarized by the SSL ethos of "Learning what is predictible". Learning with a weak world model means that it cannot model the world properly and the encoder removes the information that cannot be predicted. On the other hand, if the world model is very powerful, the representation does not need to be as abstract or semantic as it can find a way to predict representations in any situation. This means that learning a world model offers a measurable way to control the level of abstraction of the representations.

\section{Conclusion and future perspectives}

We introduced IWM, an approach to learn self-supervised visual representations with world models. With an in-depth study, we provided guidelines and key components for learning a good image world model. Conditioning the world model with the image transformation is crucial to avoid collapsing to classical SSL behavior. Using strong transformations is also key to ensure that the world model learns to model more complex behavior and be useful. Finally, enough capacity is needed for modeling complex behaviors. We showed that only a capable world model can be reused for discriminative task. This led to our predictor finetuning protocol that matches encoder finetuning at a fraction of the cost, showing that world models are versatile evaluation heads. We further adapted it to solve multiple tasks at once without losing performance. Finally, we studied how learning a world model impacts representation quality. A capable world model learns rich representations that improve performance on downstream tasks such as image classification and semantic segmentation. Additionally, learning an invariant world model led to better representations for linear evaluation. While MIM and Contrastive approaches are two ends of a spectrum in terms of representation abstraction, Image World Models allow us to interpolate between them. As such, we believe that learning image world models is a very promising framework for visual representation learning.

\section{Broader impact statement}

This paper presents work whose goal is to advance the field of Machine Learning. There are many potential societal consequences of our work, none of which we feel must be specifically highlighted here.

\bibliography{paper}
\bibliographystyle{assets/plainnat}

\newpage
\appendix
 \renewcommand{\thefigure}{S\arabic{figure}}
  \renewcommand{\thetable}{S\arabic{table}}
  \setcounter{figure}{0}    
\setcounter{table}{0}

\onecolumn
\section{Experimental details\label{app:exp-details}}
\subsection{Pretraining}
\begin{figure*}[!t]
    \centering
    \includegraphics[width=\textwidth]{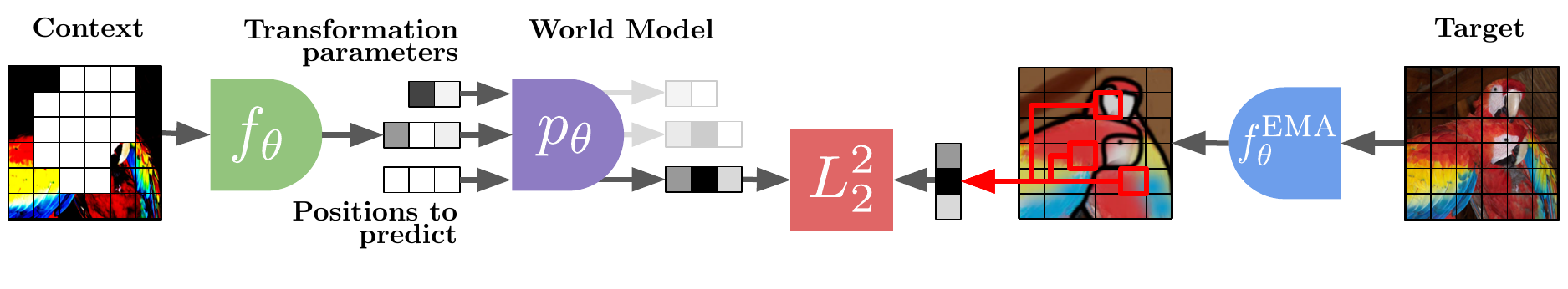} 
    \vspace{-2em}
    \caption{\textbf{IWM (Image World Model).} Starting from an image, two augmented views are produced: the source and the target. The source view is partially masked to form the context and then encoded to be used as conditioning for the world model, instantiated by the predictor. The target is encoded through an exponential moving average of the encoder, and target positions are sampled as the masked patches of the source image. Conditioned on the transformation parameters between the source and target, the encoded source image, and the positions to predict, the predictor is trained to predict the target representations.}
    \label{fig:figure-1-bis}
\end{figure*}

We provide a more detailed architecture for IWM in figure~\ref{fig:figure-1-bis}.

\textbf{Architecture and optimization.}
All of our models use a ViT-B/16 encoder trained for 300 epochs on ImageNet. We use the AdamW optimizer~\cite{loshchilov2018decoupled} with $1\times 10^{-3}$ as our learning. We further use $\beta_1 = 0.9$ and $\beta_2 = 0.999$. The learning rate follows a linear warmup  for 40 epochs and then a cosine annealing. We use an iteration per epoch scale of 1.25 for the scheduler, which stretches the scheduler and makes the training end before the end of the schedule. Not having a 0 learning rate near the end of training was found beneficial in our experiments. We use a cosine weight decay schedule which goes from $0.04$ to $0.4$.

\textbf{Source and target.}
In practice we build the source and target separately by first applying a random crop of scale between $0.3$ and $1$. We then apply a horizontal flip with probability $0.5$. We will call the resulting image $I'$.

\textbf{Target transformations.}
Starting from $I'$ we then apply a color jitter with probability $0.8$, brightness maximum strength $0.4$, contrast maximum strength $0.4$, hue maximum strength $0.1$, and saturation maximum strength $0.2$.

\textbf{Source transformations.}
Starting from $I'$ we apply a color jitter with probability $0.8$, brightness maximum strength $0.4$, contrast maximum strength $0.4$, hue maximum strength $0.1$, and saturation maximum strength $0.2$. A gaussian blur of radius between $0.1$ and $2$ is applied with probability $0.2$, solarization with probability $0.2$ and grayscale with probability $0.2$. These augmentations correspond to the ones used in BYOL~\citep{grill2020byol}.
We then generate a mask $M_x$ as the union of 4 masks of area between $0.15$ and $0.2$ of the image, with aspect ratios between $0.75$ and $1.5$. All of the patches in $M_x$ are then dropped from the source $x$.

\textbf{Predictor conditioning.}
We rely on the feature mixing strategy. Consider a mask token $m\in \mathbb{R}^d$ and $a \in \mathbb{R}^k$ a vector of $k$ scalars corresponding to augmentation parameters. We first add position embeddings to $m$ to indicate which patch of the target it needs to predict. We then concatenate $m$ and $a$ and feed them through a three layer fully-connected network with ReLU activation and dimensions $d,d,d$.  This gives us a mask token that contains information about all of the transformation. Both the geometric aspect of where to predict and details on the photometric augmentations.

\subsection{Evaluation}

For all evaluations on image classification, the augmentations applied to compute the validation accuracy are a resize to 256 followed by a 224 by 224 center crop.All hyperparameters reported are the optimal ones, chosen after careful tuning for every method.

\paragraph{Linear.}
We take inspiration from the protocol of~\cite{chen2021mocov3}. We train for 90 epochs on ImageNet. We sample random crops of images with scale between 0.08 and 1, then apply a horizontal flip with probability 0.5.\\
The features are average pooled along the sequence axis to obtain a global representation which is then fed to a linear layer.
We use a batch size of 16,384, with the LARS~\citep{you2017lars} optimizer and a learning rate of 6.4  with a warmup of 10 epochs. The learning rate then follows a cosine annealing schedule. Weight decay is set to 0 and momentum to 0.9.

\paragraph{Attentive.}
The attentive head is taken from~\cite{chen2023context}. It consists of a cross attention block where the attention is computed between an additional token the unpooled representations. This allows an adaptive pooling strategy.
We train for 90 epochs on ImageNet. We sample random crops of images with scale between 0.3 and 1, then apply a horizontal flip with probability 0.5. We also apply the same augmentations as used for the source transformations besides masking.
We use a batch size of 1024 and AdamW optimizer with a learning rate of $1\times 10^{-4}$,$\beta_1 = 0.9$, and $\beta_2 = 0.999$. It follows a cosine annealing schedule. We use a weight decay of $0.01$ kept constant during training.

\paragraph{Encoder finetuning.}
We append a linear layer to the end of the encoder as for the linear evaluation and train for 100 epochs on ImageNet. We use the same RandAugment~\citep{2020RandAug} strategy as MAE~\citep{he2021mae} as well as CutMix~\citep{yun2019cutmix} and MixUp~\citep{zhang2018mixup}. For RandAugment we use the string $\texttt{'rand-m9-mstd0.5-inc1'}$. We use random erasing with probability 0.25 in pixel mode. We use a mixup $\alpha$ of 0.8, cutmix $\alpha$ of 1 and label smoothing of 0.1.\\
For the optimization we use AdamW with a learning rate of $2\times 10^{-3}$ with 5 epochs of warmup followed by a cosine annealing schedule, weight decay of 0.005 and a batch size of 1024. We also use a drop path rate of 0.2 through the encoder and a layer wise learning rate decay of 0.65.

\paragraph{Predictor finetuning.}
When finetuning the predictor we use an attentive head on top of the predictor output. We plug the predictor on top of the teacher network and it is tasked with predicting the whole target image, with null transformation parameters. We use the same augmentation protocol as for encoder finetuning. We train for 100 epochs on ImageNet with a batch size of 1024. We use AdamW for the optimizer, a learning rate of $1\times 10^{-3}$ with a 5 epoch warmup then cosine annealing schedule. We use a weight decay of 0.1, no layer wise lr decay and a drop path rate of 0.2 through the predictor.
Importantly, if the predictor is pretrained we divide it's learning rate by 10, and keep it identical to the attentive if head if random.

\begin{figure}[!h]
    \centering
    \includegraphics[width=0.8\columnwidth]{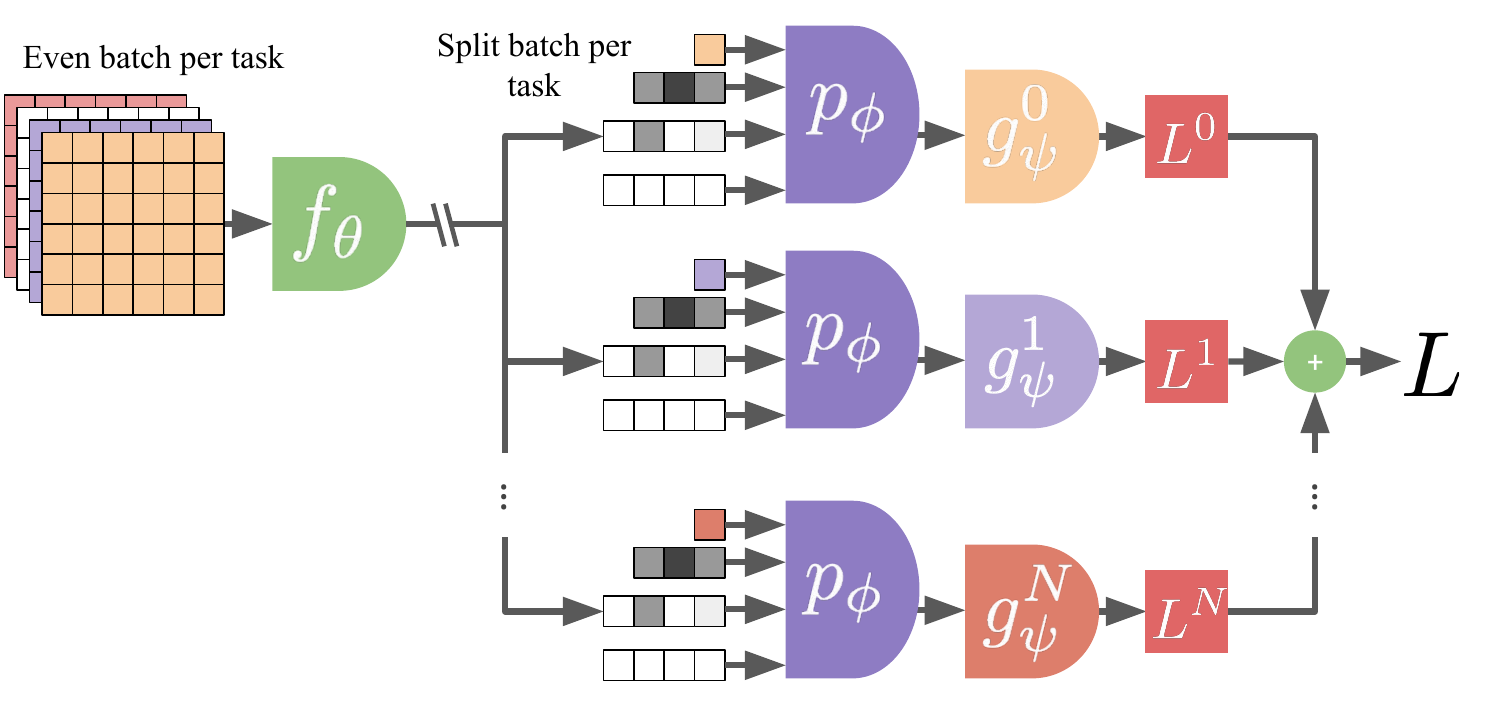}
    \caption{Multitask tuning of the predictor. We sample a batch uniformly across task which is then fed through the predictor with an additional task token, indicating which task is being solved. The predictions are then fed through a task specific head and losses are summed.}
    \label{fig:multitask}
\end{figure}

\paragraph{Multitask predictor finetuning.}
To give a clearer presentation of the protocol, we provide a graphical version of multitask predictor finetuning in figure~\ref{fig:multitask}.
For the training in itself, we follow the same protocol as for predictor finetuning but train for the equivalent of 50 ImageNet epochs. The batch size used is 512 for each task, where the batch is independently split between tasks. When training on a single task, we simply use 512 as the batch size and also train for 50 ImageNet epochs.

\paragraph{End to end finetuning.}
We follow the protocol of predictor finetuning but tweak certain parameters. First, the encoder also gets his learning rate divided by 10 like the predictor. The factors are treated separately and ablated for all methods. We use a 0.9 layer decay across the combination of predictor and encoder. The learning rate used is $2\times 10^{-3}$ and all other parameters are identical to predictor finetuning.

\paragraph{Segmentation.} \label{app:seg}

We give here details about our protocol for semantic segmentation evaluations. We use the MMSegmentation library~\cite{mmseg2020}. We fine-tune our pretrained models (either encoder only, predictor only, or end-to-end) with an UperNet head~\cite{xiao2018upernet} on the ADE20k semantic segmentation dataset~\cite{zhou2019ade20k} for 160k iterations and report the validation mIoU. We concatenate the last 4 layers of the predictor, or encoder for encoder only finetuning, and feed the result to the segmentation head. At training time we resize the images at the pretraining resolution. At testing time we do not resize the images and interpolate the positional embeddings to the original resolution. For all setups and methods we pick the best run among several learning rate values: $1e-5$, $2e-5$ and $3e-5$. We use a weight decay of $0.01$ and a linear learning rate decay schedule.

\section{Complete finetuning results}
\begin{table*}[!h]
\centering
    \caption{Complete results of tables~\ref{tab:full-results} and~\ref{tab:results_peak}.}
    \label{tab:full-results-appdx}
\begin{tabular}{lccccc}
\toprule
\multirow{2}{*}{Method} & \multirow{2}{*}{Epochs} & \multicolumn{1}{c}{No predictor} & \multicolumn{2}{c}{Frozen encoder, tuned predictor}  & \multirow{2}{*}{End to end}  \\
\cmidrule(l){3-3} \cmidrule(l){4-5}
& & Encoder & Random Init. & Pretrained &   \\
\midrule
DINO & 1600 & 82.8 & 82.0 & N/A & 82.1 \\
MOCOv3 & 300 & 83.2 & 56.4 & N/A & 79.4  \\
iBOT& 1600 & \textbf{84.0} & \textbf{83.0} & N/A & 82.8 \\
\multirow{ 3}{*}{MAE} & 300  & 82.7 & 82.4 & 82.7 \textcolor{Red}{(+0.3)} & 82.3  \\
& 600  & 83.2 & 82.8 & 83.0 \textcolor{Red}{(+0.2)} & 83.1  \\
& 1600  & 83.6 &  \textbf{83.0} & 83.1 \textcolor{Red}{(+0.1)} & 83.3  \\
\multirow{ 1}{*}{I-JEPA} & 300 & 83.0 & 79.1 & 80.0 \textcolor{orange}{(+0.9)} & 82.0   \\
 \midrule
\multirow{ 1}{*}{$\text{IWM}_{12,384}^\text{Inv}$} & 300  & 83.3 & 80.5 & 81.3 \textcolor{orange}{(+0.8)} & 82.7  \\
$\text{IWM}_{12,384}^\text{Equi}$ & 300 & 82.7 & 81.3 & 82.7 \textcolor{Green}{(+1.4)} & 83.3  \\
$\text{IWM}_{18,384}^\text{Equi}$ & 300 & 82.9 & 81.5 & \textbf{83.3} \textcolor{Green}{(+1.8)} &  \textbf{84.4} \\

\bottomrule
\end{tabular}

\end{table*}

We provide complete results for table~\ref{tab:full-results} and table~\ref{tab:results_peak} in table~\ref{tab:full-results-appdx}. Some interesting behaviors are $\text{IWM}_{12,384}^\text{Equi}$ and MoCov3 in predictor finetuning. For $\text{IWM}_{12,384}^\text{Equi}$, we see the same behavior as $\text{IWM}_{18,384}^\text{Equi}$ but with slightly lower performance. This is consistent across all evaluations. Yet, even when accounting for scale of the predictor to compare with I-JEPA and $\text{IWM}_{12,384}^\text{Inv}$, all of our previous conclusions still hold.
For MoCov3, it was the only method which did not perform well when attaching a random predictor to it. While we do not have conclusive evidence, we hypothesize that it is related to the low norm of its output. Adding a normalization between the encoder and predictor did not help.

\clearpage
\section{Impact of data augmentation \label{app:data-aug}}

\begin{table}[!h]
\centering
    \caption{Impact of data augmentation strategy on IWM's performance. In all settings, destructive augmentations are never applied to the target.}
    \label{tab:full-aug-results}
\resizebox{\columnwidth}{!}{\begin{tabular}{lcccccccc|cccc}
\toprule
\multirow{2}{*}{Predictor} & \multicolumn{4}{c}{Strengths}& \multicolumn{4}{c}{Probabilities} & \multicolumn{4}{c}{Performance}\\
\cmidrule(l){2-5} \cmidrule(l){6-9} \cmidrule(l){10-13}
&  Bright. & Contrast & Sat. & Hue & Jitter & Blur & Gray. & Solarize & MRR & Linear & Attentive & Pred. ft. \\
\midrule
\rowcolor{Gray}
\multirow{6}{*}{$\text{IWM}_{12,384}$} & 0.4 & 0.4 & 0.2 & 0.1 & 0.8 & 0.2 & 0.2 & 0.1 & 0.09 & 74.5 & 77.0 & 81.3 \\
&  0.4 & 0.4 & 0.2 & 0.1 & 0.8 &  &  &  & 0.11 & 72.1 & 76.5 & 80.5 \\
& 0.4 & 0.4 & 0.2 & 0.1 & 0.8 & 0.4 & 0.4 & 0.2 & 0.22 & 71.0 & 74.9 &80.9 \\
&  0.5 & 0.5 & 0.4 & 0.2 & 0.8 & 0.2 & 0.2 & 0.1 & 0.81 & 69.3 & 75.5 & 82.7  \\
&  0.5 & 0.5 & 0.4 & 0.2 & 0.8 &  &  &  & 0.07 & 73.3 & 76.3 & 80.1 \\
&   &  &  &  &  & 0.2 & 0.2 & 0.1 & 0.02 & 72.9 & 76.3 & 80.7 \\
\midrule
\rowcolor{Gray}
\multirow{6}{*}{$\text{IWM}_{18,384}$} & 0.4 & 0.4 & 0.2 & 0.1 & 0.8 & 0.2 & 0.2 & 0.1 & 0.79 & 67.5 & 75.1 & 83.3 \\
&  0.4 & 0.4 & 0.2 & 0.1 & 0.8 &  &  &  & 0.25 & 70.1 & 74.8 & 81.4 \\
& 0.4 & 0.4 & 0.2 & 0.1 & 0.8 & 0.4 & 0.4 & 0.2 & 0.85 & 56.1 & 74.5 & 83.1 \\
&  0.5 & 0.5 & 0.4 & 0.2 & 0.8 & 0.2 & 0.2 & 0.1 & 0.85 & 34.3 & 71.0 & 81.7 \\
&  0.5 & 0.5 & 0.4 & 0.2 & 0.8 &  &  &  & 0.83 & 69.2 & 75.8 & 83.3 \\
&   &  &  &  &  & 0.2 & 0.2 & 0.1 & 0.02 & 70.9 & 74.8 & 81.5 \\
\bottomrule
\end{tabular}}
\end{table}

We study in table~\ref{tab:full-aug-results} the impact of augmentations used during pretraining, along with the depth of the predictor. We notice that depth is a deciding factor in the quality of the learned world model, where 4 out 5 scenarios with color are able to achieve color equivariance for the 18 layer predictor, compared to only 1 for the 12 layer predictor. The strength of the augmentations also plays a role and too weak augmentations do not lead to an equivariant model.

\paragraph{On the asymmetry of augmentations.}
The asymmetry of augmentations is both a conceptual choice, to make the augmentations used more similar to contrastive approaches, but also a practical one. When learning an invariant world model with symmetric augmentations we noticed a drop in performance of 2 points on ImageNet in attentive probing and 1.5 points on linear probing. While this drop is not catastrophic, it is sufficient to recommend using asymmetric augmentations. As the depth of the predictor decreases, we expect this gap to widen.\\
On the other hand, when looking at an equivariant predictor, we did not notice any notable change in performance. This suggests that learning world models can also help improve stability over the choice of augmentations. The predictor does not have to be designed by keeping in mind which information may get removed but only by whether or not it can apply the transformation.

\section{Impact of the prediction task on predictor finetuning performance\label{app:pred-task}}

In order to use predictor finetuning to solve downstream tasks, we need to apply a prediction task. We aim at giving a more combinatorial view of table~\ref{tab:pred-ft-ablation} in this appendix.

\begin{table}[!h]
\centering
    \caption{Predictor finetuning performance which different prediction tasks.}
    \label{tab:full-pred-ablation}
\begin{tabular}{lcccc}
\toprule
Method & Null latents & On teacher & Pred only one token & Accuracy  \\
\midrule
\rowcolor{Gray}
\multirow{8}{*}{$\text{IWM}_{12,384}$} & \checkmark & \checkmark &  & 83.3 \\
& \checkmark & \checkmark & \checkmark & 82.8 \\
& \checkmark &  &  & 83.1 \\
& \checkmark &  & \checkmark & 82.6 \\
&  & \checkmark &  & 83.2 \\
&  & \checkmark & \checkmark & 82.8 \\
&  &  &  & 82.9 \\
&  &  & \checkmark & 82.9 \\
\bottomrule
\end{tabular}
\end{table}

We can see in table~\ref{tab:full-pred-ablation} that the conclusions drawn from table~\ref{tab:pred-ft-ablation} still hold over a larger setting. Notably, using null latents is more flexible while not changing performance, using the teacher always gives a small boost in performance, and predicting only one token lowers performane by roughly half a point.

\section{Scaling to larger models\label{app:large}}

In order to scale to larger models, such as a ViT-L/16 encoder, multiple challenges need to be overcome. Notably, both the depth and the width of the predictor must be scaled in order to increase the number of parameters of the predictor to a suitable number. Scaling the width can lead to instabilities and hyperparameters such as the EMA schedule become more important. We noticed that a ratio of predictor weights/encoder weights of around 0.3 is suitable to learn a good world model.

\begin{table}[!h]
\centering
    \caption{With a ViT-L/16 encoder, we observe a similar trend as with the base model. Significant gains are observed with a good world model, allowing it to surpass encoder finetuning.}
    \label{fig:results-large}
\begin{tabular}{lcccc}
\toprule
Method & Epochs & Encoder & Predictor & End to end \\
\midrule
I-JEPA & 300 & 84.1 & 79.9 &  \\
$\text{IWM}_{18,384}^\text{Inv}$ & 300  & 84.3 & 81.5  & \\
$\text{IWM}_{36,512}^\text{Equi}$ & 300  & 83.7 & 85.0  & 85.4  \\
\bottomrule
\end{tabular}
\end{table}

We study in table~\ref{fig:results-large} the performance when scaling to a larger ViT-L/16. We see that the observation we made with the smaller ViT-B/16 still hold. The invariant model is the best on encoder finetuning, and predictor finetuning improves the performance significantly. Here again, end-to-end finetuning leads to performance gains.

\section{Evaluation on downstream datasets beyond ImageNet\label{app:ood}}

We evaluated I-JEPA and IWM on iNaturalist18~\citep{vanhorni2018naturalist}, SUN397~\citep{xiao2010sun} and Places205~\citep{zhou2014places} using attentive probing. We train our models for 50 epochs on iNaturalist18, 12 for Places205 and 28 for SUN397.

\begin{table}[!h]
\centering
    \caption{When evaluating with attentive probing on downstream task, being equivariant improves performance across the board. All methods use ViT-B/16 encoders and were pretrained for 300 epochs on ImageNet}
    \label{tab:ood}
\begin{tabular}{lcccc}
\toprule
Method  & ImageNet & iNat18 & SUN397 & Places205   \\
\midrule
MAE  & 73.5 & 50.1 & 70.2 & 60.3 \\
I-JEPA & 75.0 & 50.4 & 69.2 & 58.3 \\
$\text{IWM}_{12,384}^\text{Inv}$ & \textbf{77.0} & 51.6 & 71.0 & 59.4 \\
$\text{IWM}_{18,384}^\text{Equi}$ & 75.1 & \textbf{54.2} & \textbf{71.7} & \textbf{60.5} \\

\bottomrule
\end{tabular}

\end{table}

As we can see in table~\ref{tab:ood}, IWM consistently improves over I-JEPA and MAE when pretraining all methods for 300 epochs. We notice that while $\text{IWM}_{18,384}^\text{Equi}$ is not the top performing model on ImageNet, it significantly outperforms it's invariant counterpart, with gains of 2.6 points on iNaturalist, 0.7 points on SUN397 and 1.1 point on Places205. This suggests that while the richness of the representation of an equivariant model is not optimal for in domain performance, it helps improve generalisation to downstream tasks.

\clearpage
\section{Visualizing representation differences between invariant and equivariant behavior \label{sec:repr-visualization}}

\begin{figure}[!h]
  \centering
  \begin{subfigure}{0.3\linewidth}
  \centering
    \includegraphics[width=1\textwidth]{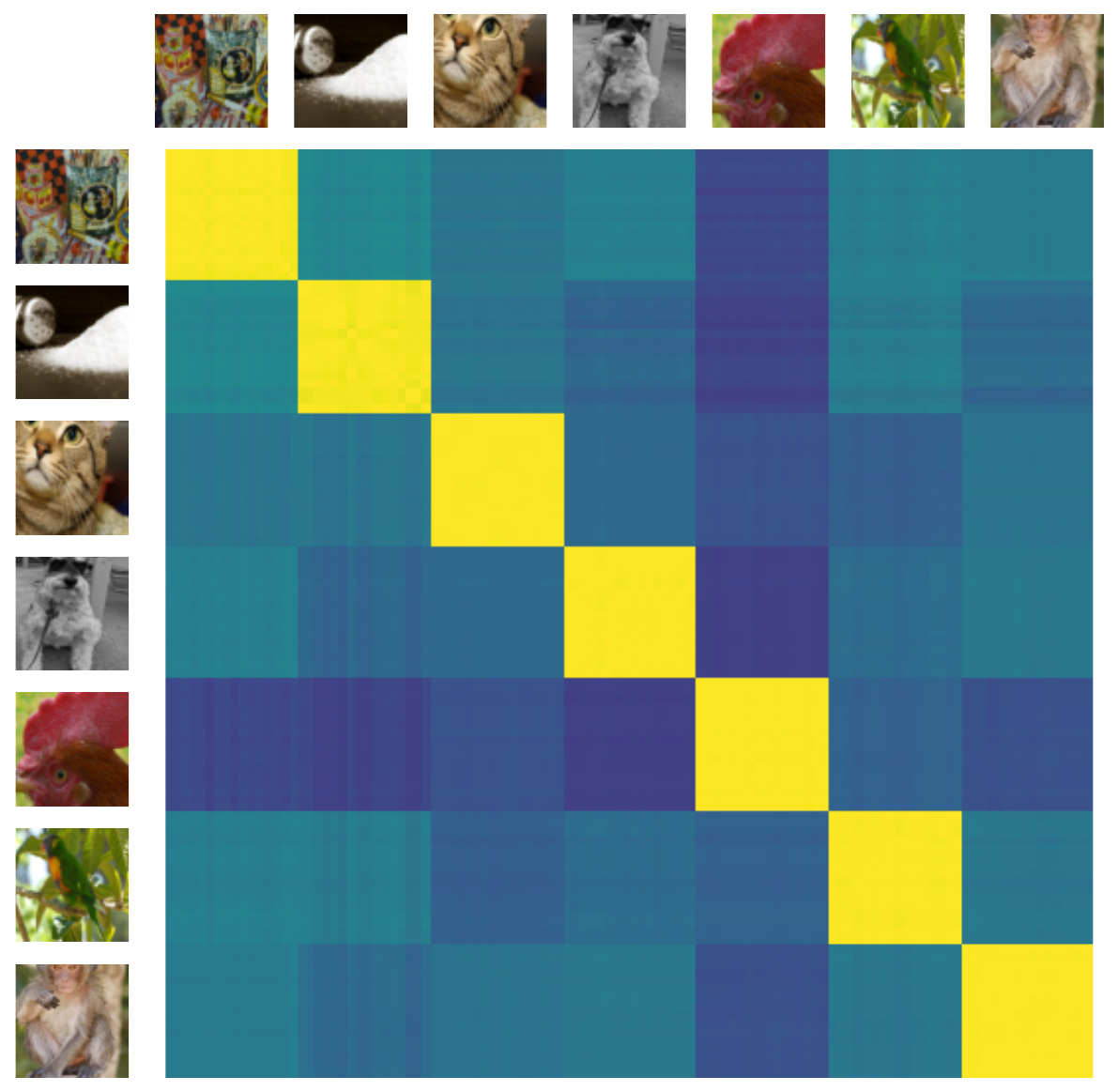}
    \label{fig:subfig1}
    \caption{$\text{IWM}_{12,384}^\text{Inv}$}
  \end{subfigure}
  \begin{subfigure}{0.3\linewidth}
  \centering
    \includegraphics[width=1\textwidth]{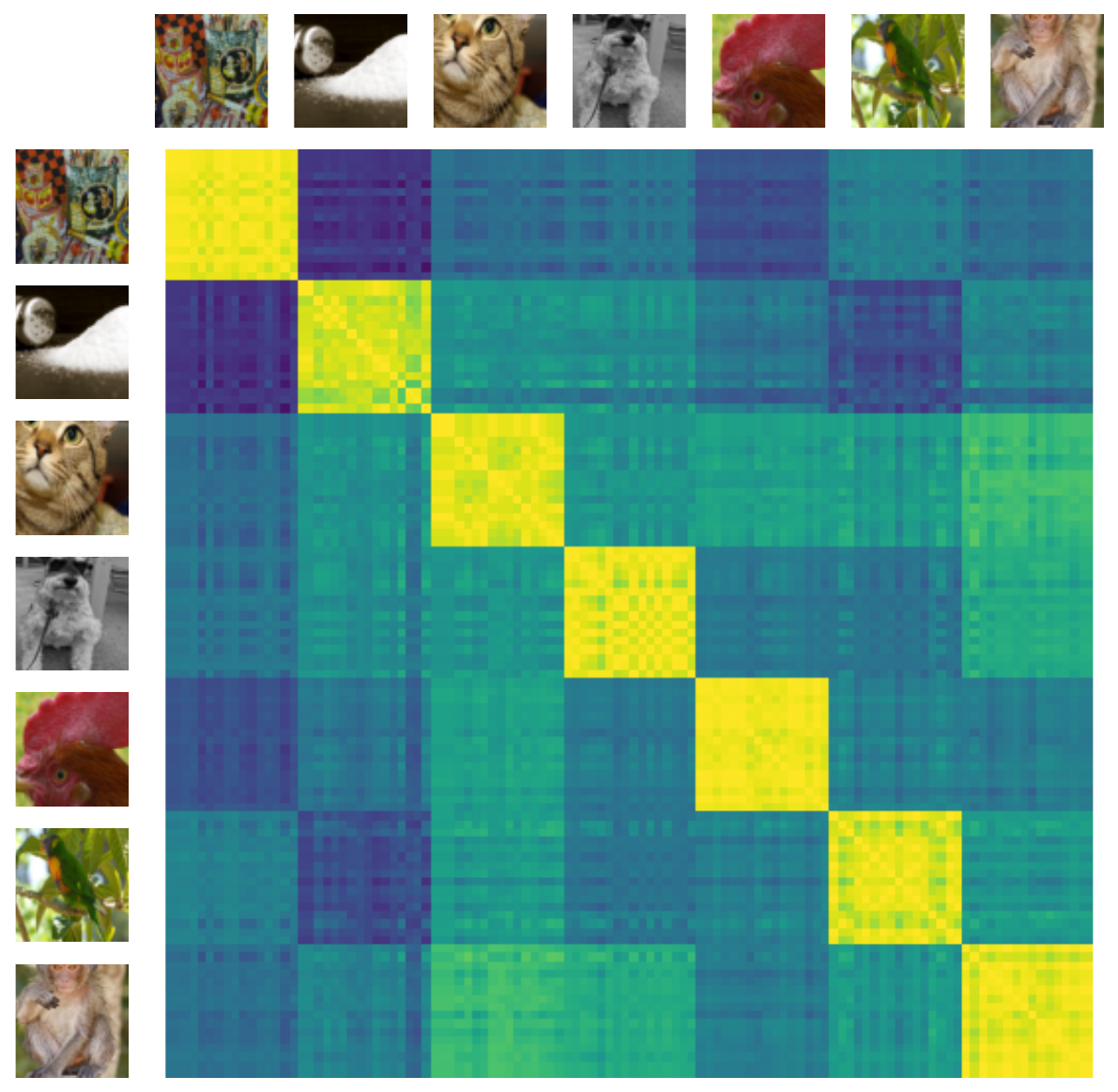}
    \label{fig:subfig2}
    \caption{$\text{IWM}_{18,384}^\text{Equi}$}
  \end{subfigure}
  \begin{subfigure}{0.3\linewidth}
  \centering
    \includegraphics[width=1\textwidth]{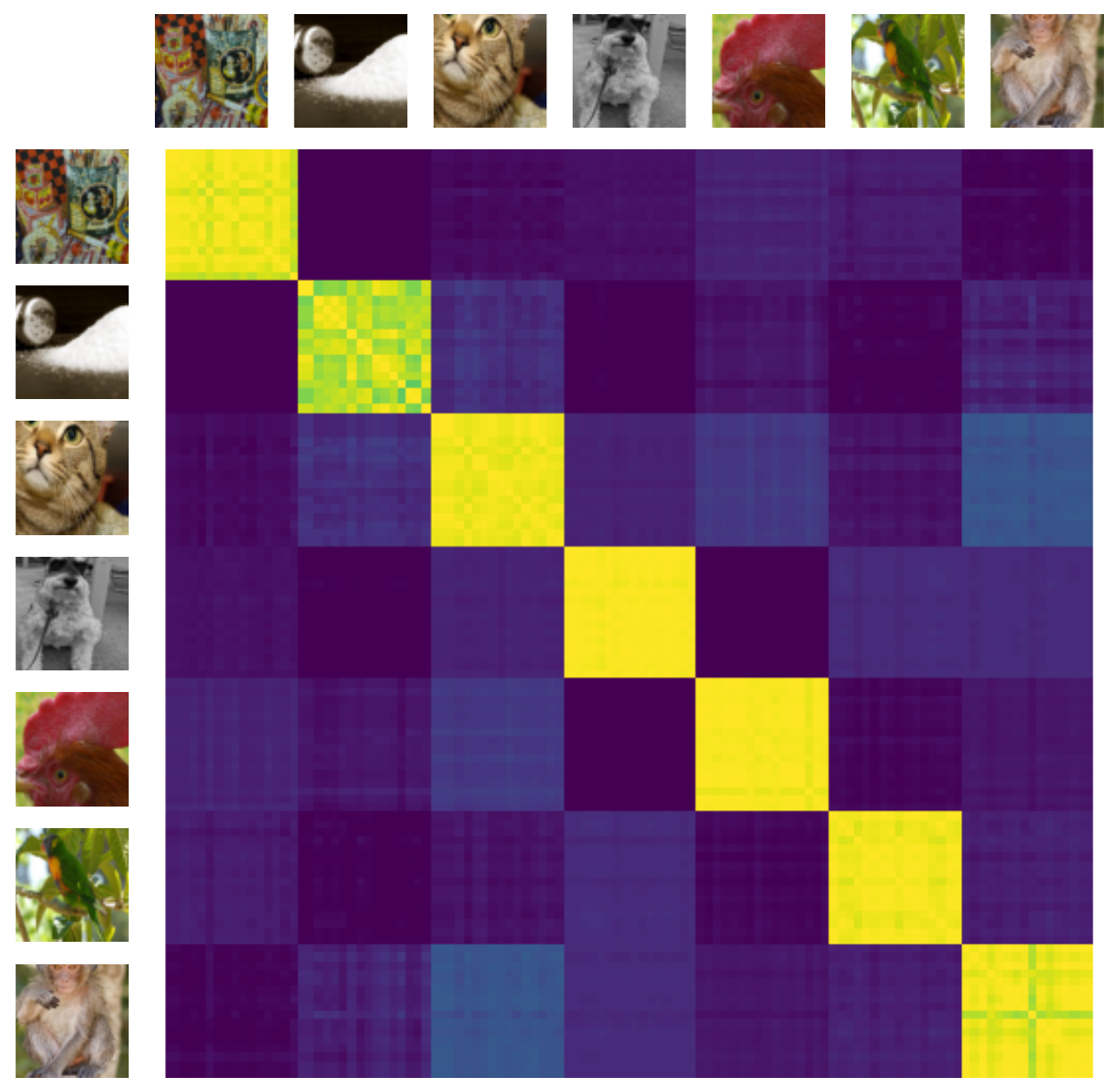}
    \label{fig:subfig3}
    \caption{IJEPA}
  \end{subfigure}
  \caption{Difference in embedding space between invariant and equivariant behaviours. Each image is augmented 16 times and we compute the similarity matrix between all images. The yellow regions indicate high similarities between samples originating from the same image. We can see more variations in the equivariant model, or in I-JEPA where invariance is not enforced. This suggests that augmentations influence the representation more in these models.}
  \label{fig:inv-vs-equi}
\end{figure}

As we see in figure~\ref{fig:inv-vs-equi}, the invariant model collapses augmented views to very similar embeddings, as shown by the high similarity in the diagonal blocks. On the other hand the equivariant model shows more variation, which shows that augmentation information is more present in the representation. Interestingly, I-JEPA has a behaviour in between because it was not trained to be either invariant or equivariant. I-JEPA has no force controlling how information is kept or removed from the representation.

\section{On the meaning and role of invariance in Self-Supervised learning}

One of the key component of the success of self-supervised learning is augmentation invariance~\cite{chen2020simple}. We can say that we have learned invariant representations if $\forall a, \; f_\theta(x) = f_\theta(\mathcal{T}(a,x))$. However there are many scenarios that satisfy this property. The two main ones that we are interested in are:
\begin{itemize}
    \item Any augmented view leads to the same information as the clean image
    \item The encoder removes the information related to the transformation
\end{itemize}

In the first case, the representations still contain all of the information about the input, whereas in the second we are removing information that can be deemed superfluous. In the case of contrastive methods, the focus is usually on removing information. Indeed if an image and its grayscale version are made to have the same representation, the encoder must remove color information. This is one of the key drivers of performance of such methods. By removing information until only the semantics of the image remains, the representations will be easy to leverage for a task such as classification.

We can thus wonder if the first invariance scenario also leads to improved performance, and if we can even leverage it. As we have demonstrated how IWM is able to preserve information, and we have a predictor that can apply transformations, we can marginalize over augmentations to create invariant representations in an efficient way. Here, we do not need to apply the encoder on all augmented views, but can directly use the predictor which is more compute efficient. If we consider a set of randomly sampled augmentations $A$ such that $\text{card}(A) = N $ we can compute an invariant representation as
\begin{equation*}
    z_x^{\text{Inv}} = \frac{1}{N} \sum_{i=1}^N p_\phi\left( f_\theta(x),A_i,m_{A_i} \right)
\end{equation*}

We can then visualize which image has representation most similar to $ z_x^{\text{Inv}}$ and see if using $ z_x^{\text{Inv}}$ improves performance on classification task.

\begin{figure}[!h]
    \centering
    \includegraphics[width=\textwidth]{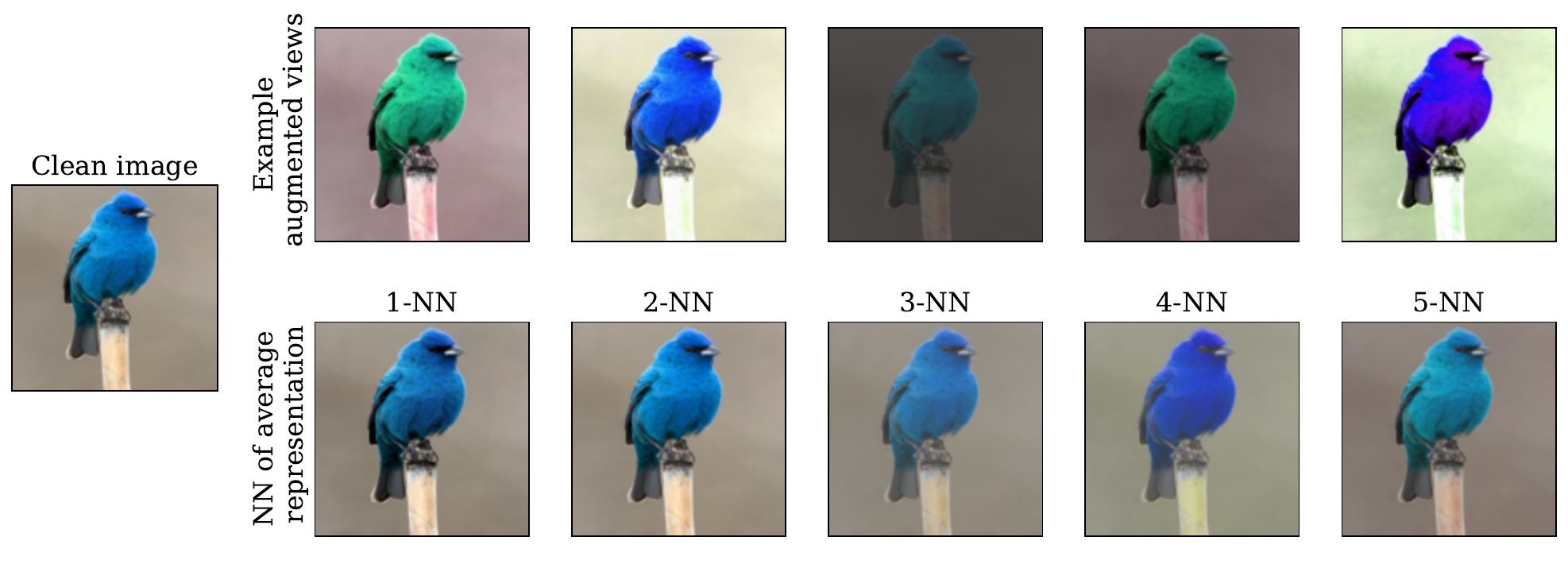}
    \caption{Retrieval of invariant representations computed using 256 augmentations in latent space. In the top row we visualize some of the corresponding image and on the bottom the nearest neighbours of the invariant representation. We can notice that the nearest neighbour is the original non-augmented image, followed by images with small transformations.}
    \label{fig:retrieval-identity-bird}
\end{figure}

As we can see in figure~\ref{fig:retrieval-identity-bird},  the images that have representations which are most similar with  $z_x^{\text{Inv}}$ are the clean image and images with small transformations. We also know that our encoder preserves augmentation related information and is thus not invaraint to transformations. Combining these two facts tells us that the marginalizaiton process creates a clean representations, akin to the first kind of invariance.

\begin{table}[!h]
\centering
    \caption{Linear evaluation on marginalized representations. Using more augmented prediction to create an invariant representation does not improve performance. }
    \label{tab:marginalisation}
\begin{tabular}{lccccc}
\toprule
Number of predictions ($N$) & 1 (default) & 8 & 16 & 32 & 128   \\
\midrule
ImageNet Top-1 accuracy (\%) & 64.5 & 64.3 & 64.6 & 64.6 & 64.4 \\
\bottomrule
\end{tabular}
\end{table}

However, when looking at table~\ref{tab:marginalisation} we can see that no performance gain is present when using invariant representations obtained by marginalizing over predictions. This is true even with 128 augmented views, which already increases the compute budget by a factor of around 64. As such, using invariant representations that preserve the content of the image is not necessarily beneficial for downstream evaluation.

Overall, the key to the success of augmentation invariance in contrastive learning is not just in building invariant representations, but in the way that the representations are invariant. Building invaraince by removal of information has been shown to be very effective~\citep{chen2020simple}, whereas we see here that invariance by always predicting the representation of the clean image is not necessarily helpful. This does not mean that equivariant representations cannot build invariances that are useful from downstream tasks, as the contrary was shown in~\cite{chavhan2023amortised}, but that we have to be careful in how we create invariant representations.

\clearpage

\section{Additional qualitative evaluations of the world model \label{app:more-qualitative}}

\begin{figure}[!h]
    \centering
    \includegraphics[width=\textwidth]{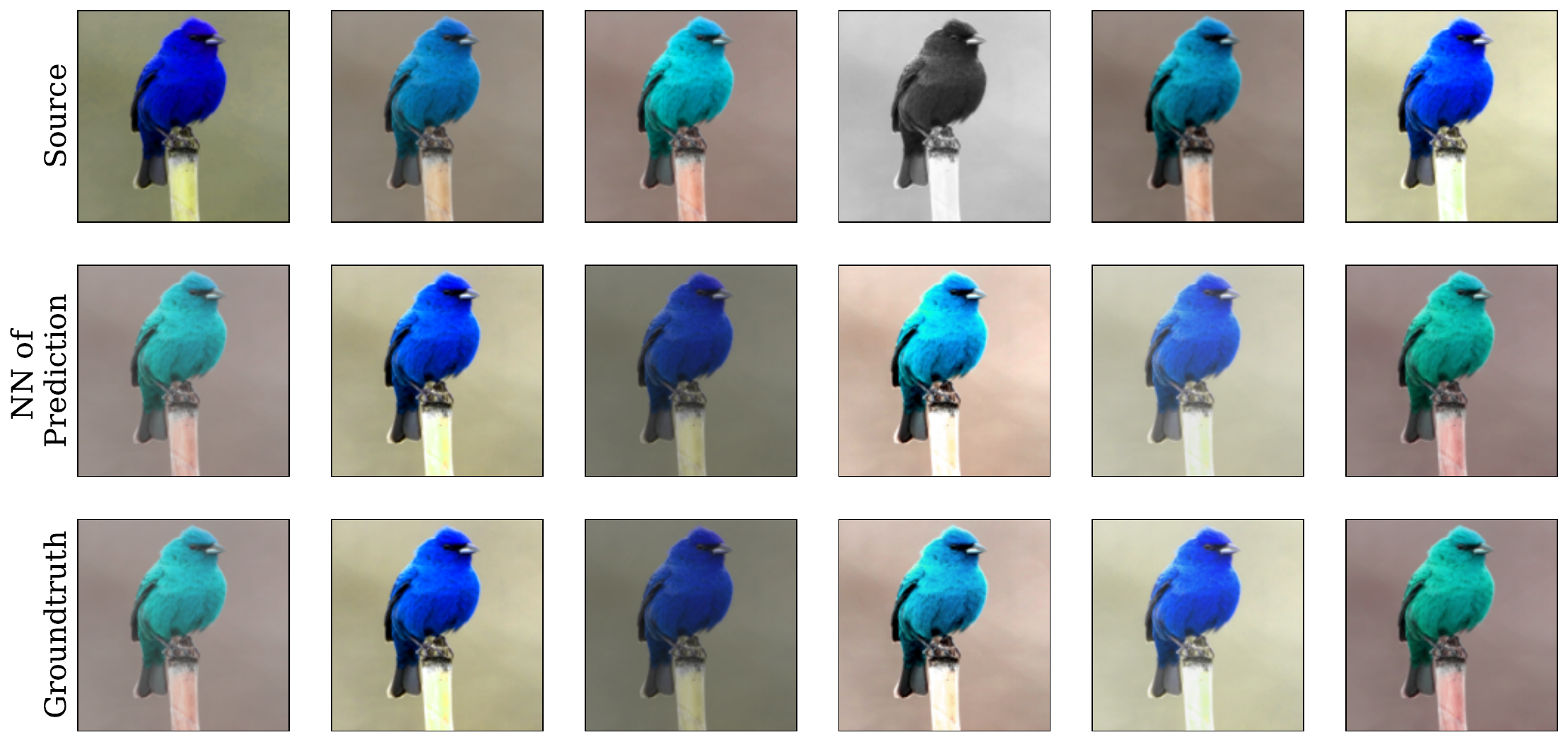}

    \caption{Randomly selected retrieval samples of our world model. For each image, we generate 256 augmented views and apply transformations in latent space. We then retrieve the nearest neighbor of the prediction and visualize whether it is close to the groundtruth or not. The learned world model performs well in most settings but has some inaccuracies with inverting grayscale.}
    \label{fig:enter-label}
\end{figure}
\begin{figure}[!h]
    \centering
    \includegraphics[width=\textwidth]{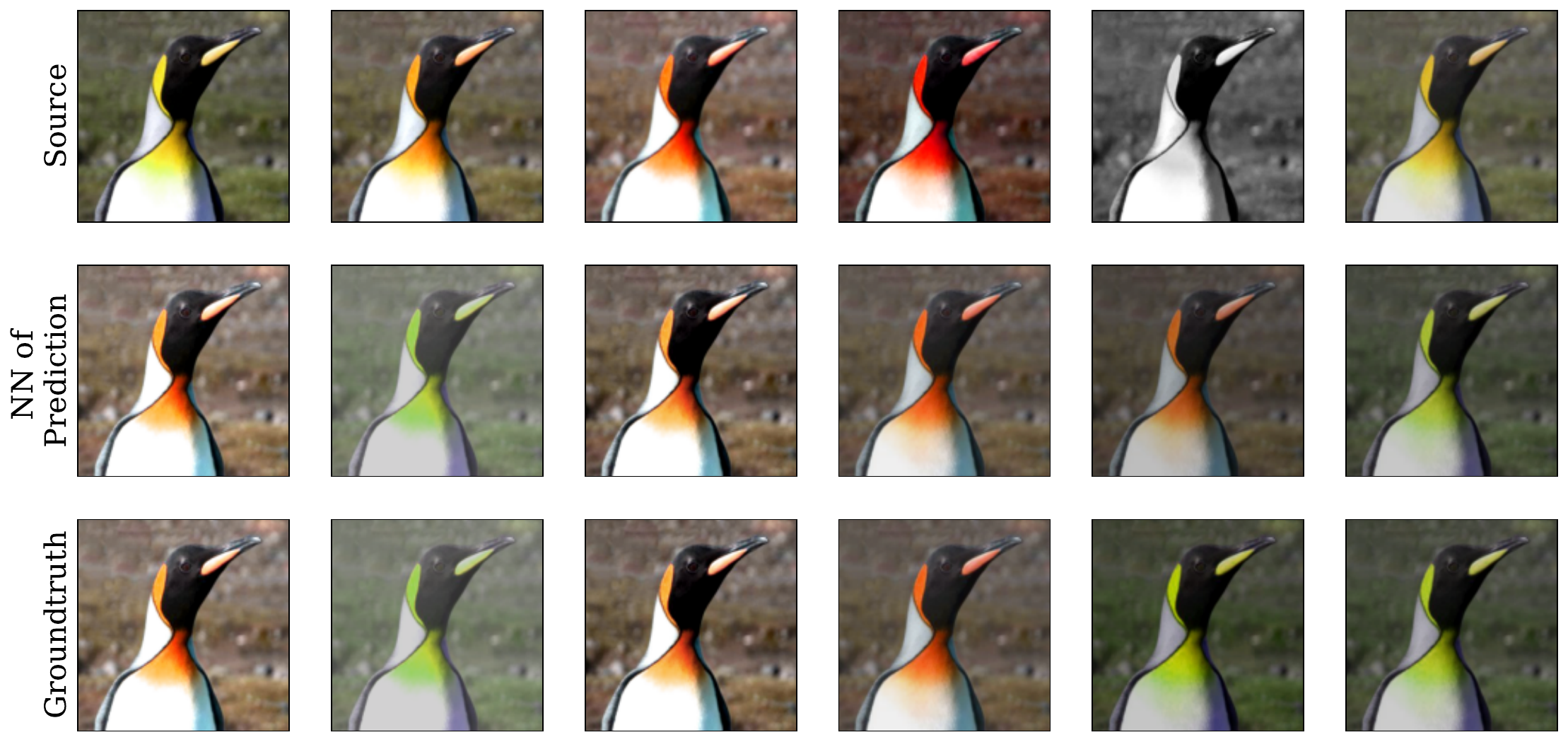}

    \caption{Randomly selected retrieval samples of our world model. For each image, we generate 256 augmented views and apply transformations in latent space. We then retrieve the nearest neighbor of the prediction and visualize whether it is close to the groundtruth or not. The learned world model performs well in most settings but has some inaccuracies with inverting grayscale.}
    \label{fig:enter-label}
\end{figure}
\begin{figure}[!h]
    \centering
    \includegraphics[width=\textwidth]{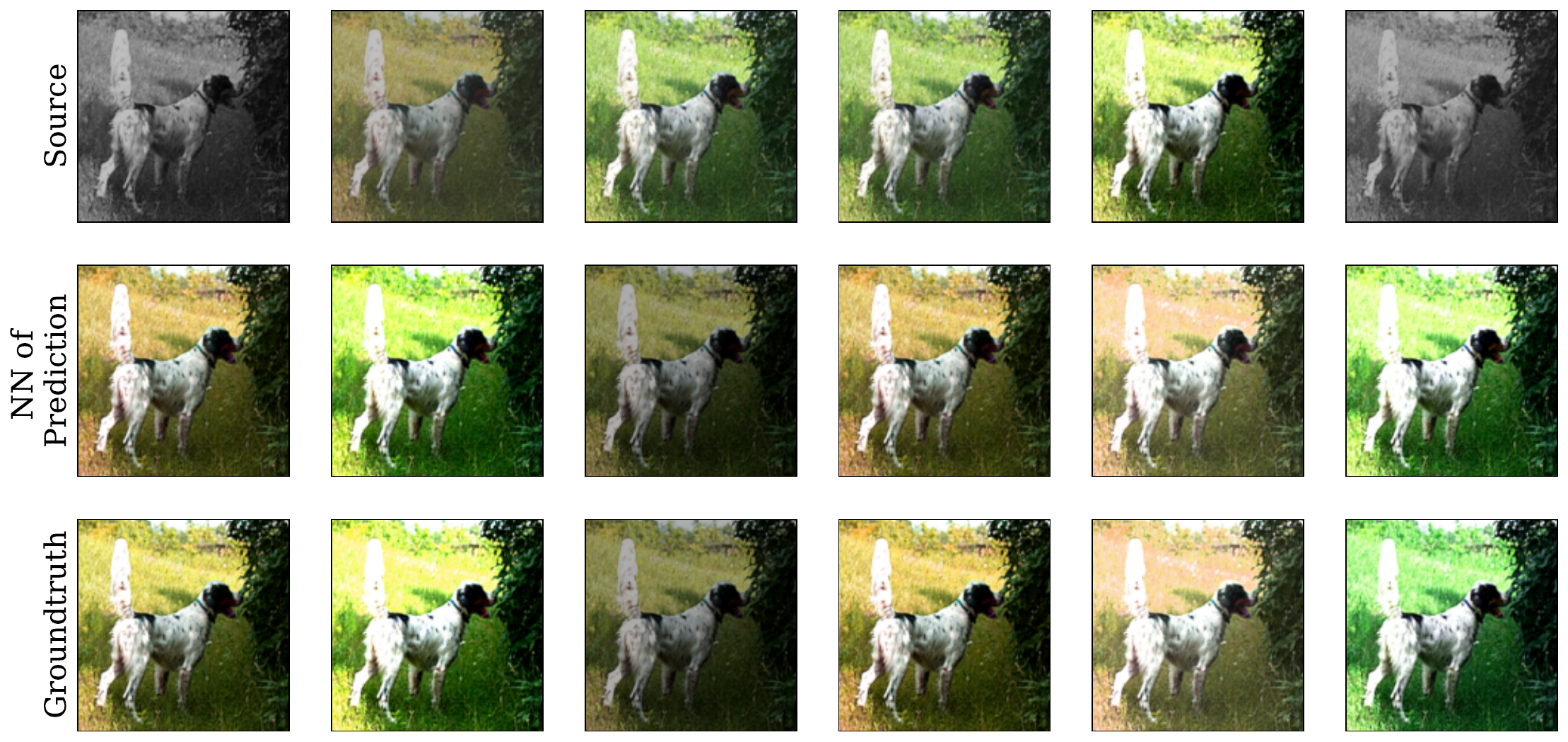}

    \caption{Randomly selected retrieval samples of our world model. For each image, we generate 256 augmented views and apply transformations in latent space. We then retrieve the nearest neighbor of the prediction and visualize whether it is close to the groundtruth or not. The learned world model performs well in most settings but has some inaccuracies with inverting grayscale.}
    \label{fig:enter-label}
\end{figure}
\begin{figure}[!h]
    \centering
    \includegraphics[width=\textwidth]{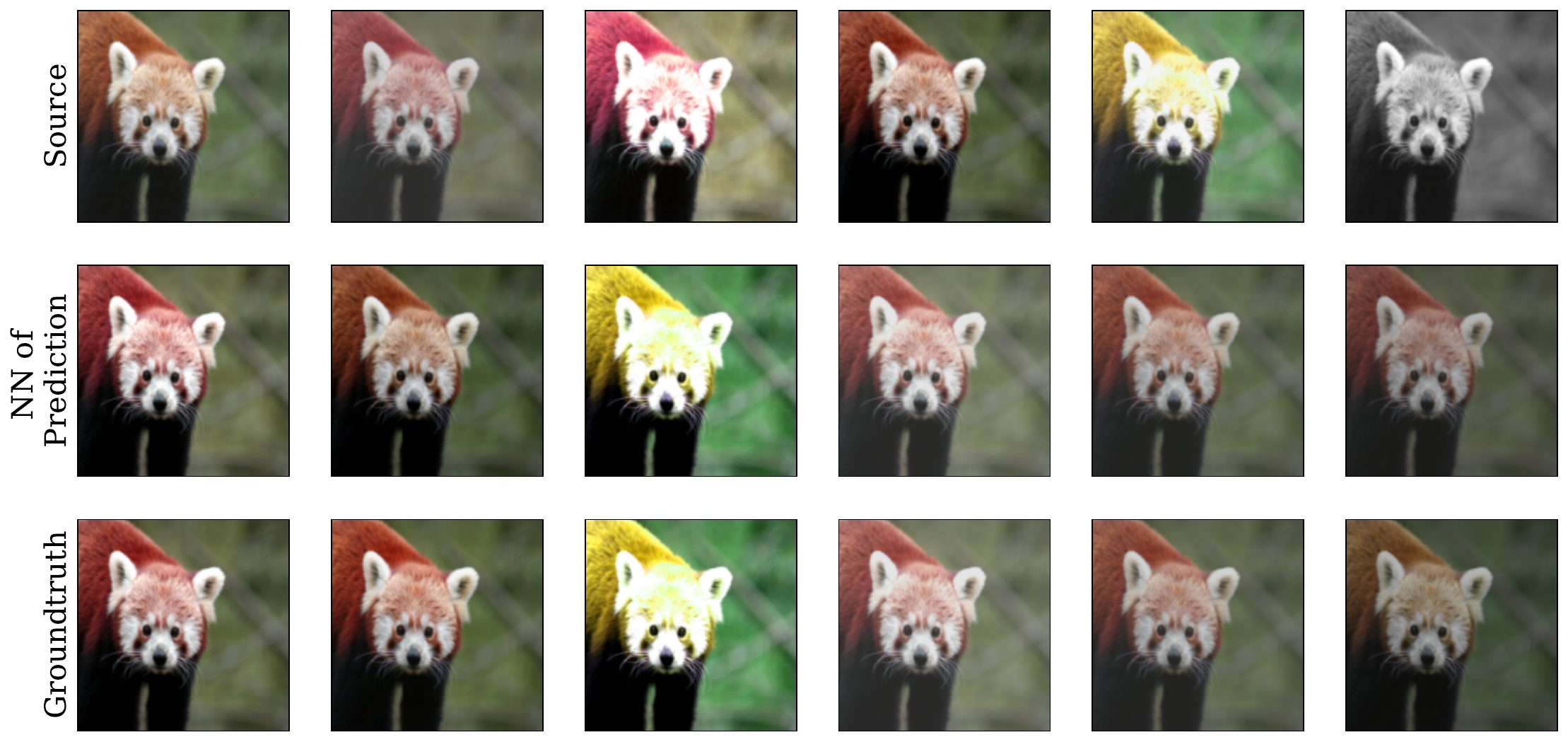}

    \caption{Randomly selected retrieval samples of our world model. For each image, we generate 256 augmented views and apply transformations in latent space. We then retrieve the nearest neighbor of the prediction and visualize whether it is close to the groundtruth or not. The learned world model performs well in most settings but has some inaccuracies with inverting grayscale.}
    \label{fig:enter-label}
\end{figure}

\begin{figure}[!h]
    \centering
    \includegraphics[width=\textwidth]{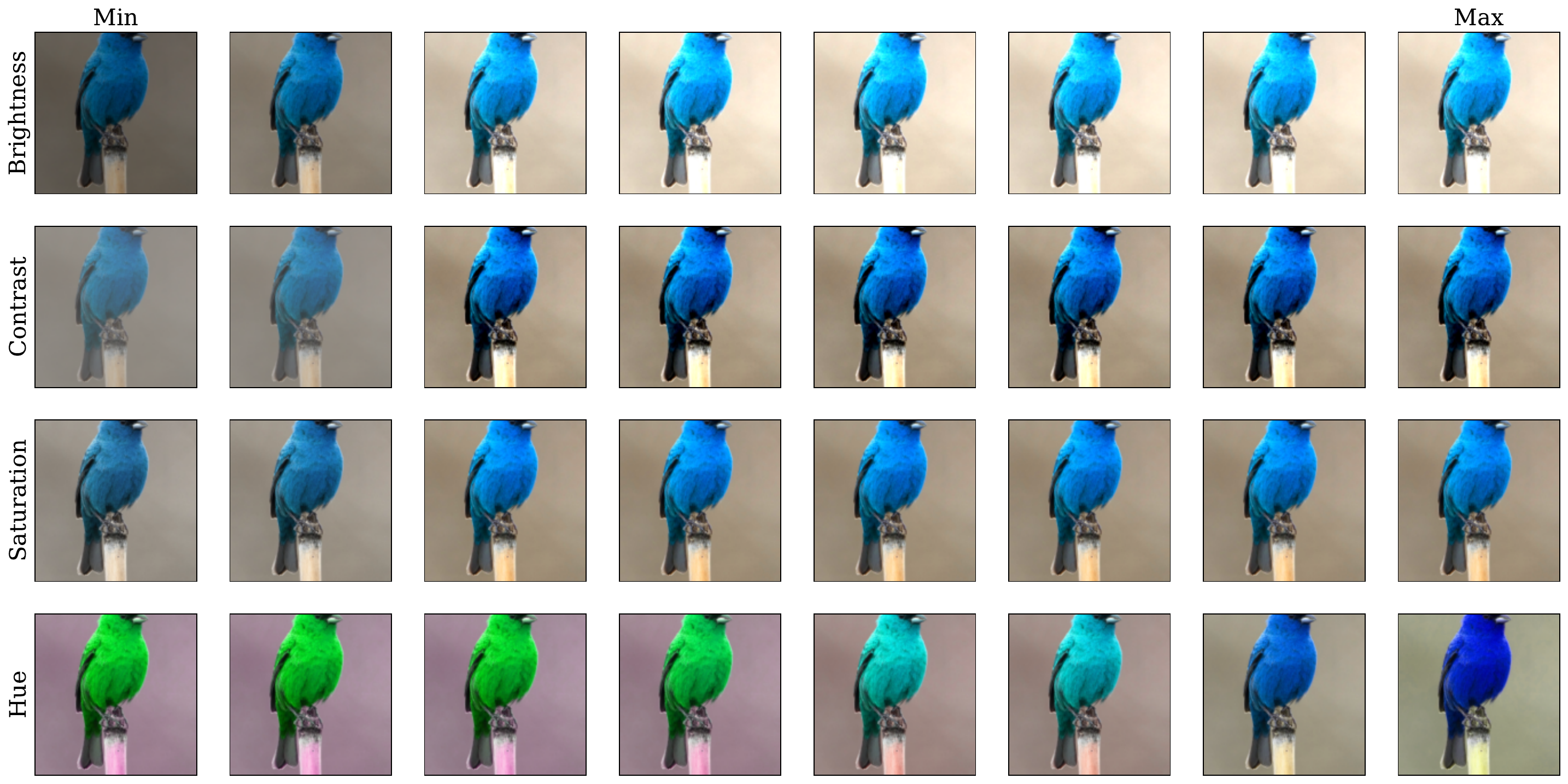}
    \caption{Application of the world model on precise transformations. For each parameter, we vary its value on a grid to see whether the model is able to predict small changes. The model is able to show the gradient of transformations, highlighting again the capabilities of the world model. We can still notice some imperfections however, as the model was only trained on combinations of augmentations. To make changes more visible, we used a model trained with a strong color jitter for this figure.}
    \label{fig:enter-label}
\end{figure}
\begin{figure}[!h]
    \centering
    \includegraphics[width=\textwidth]{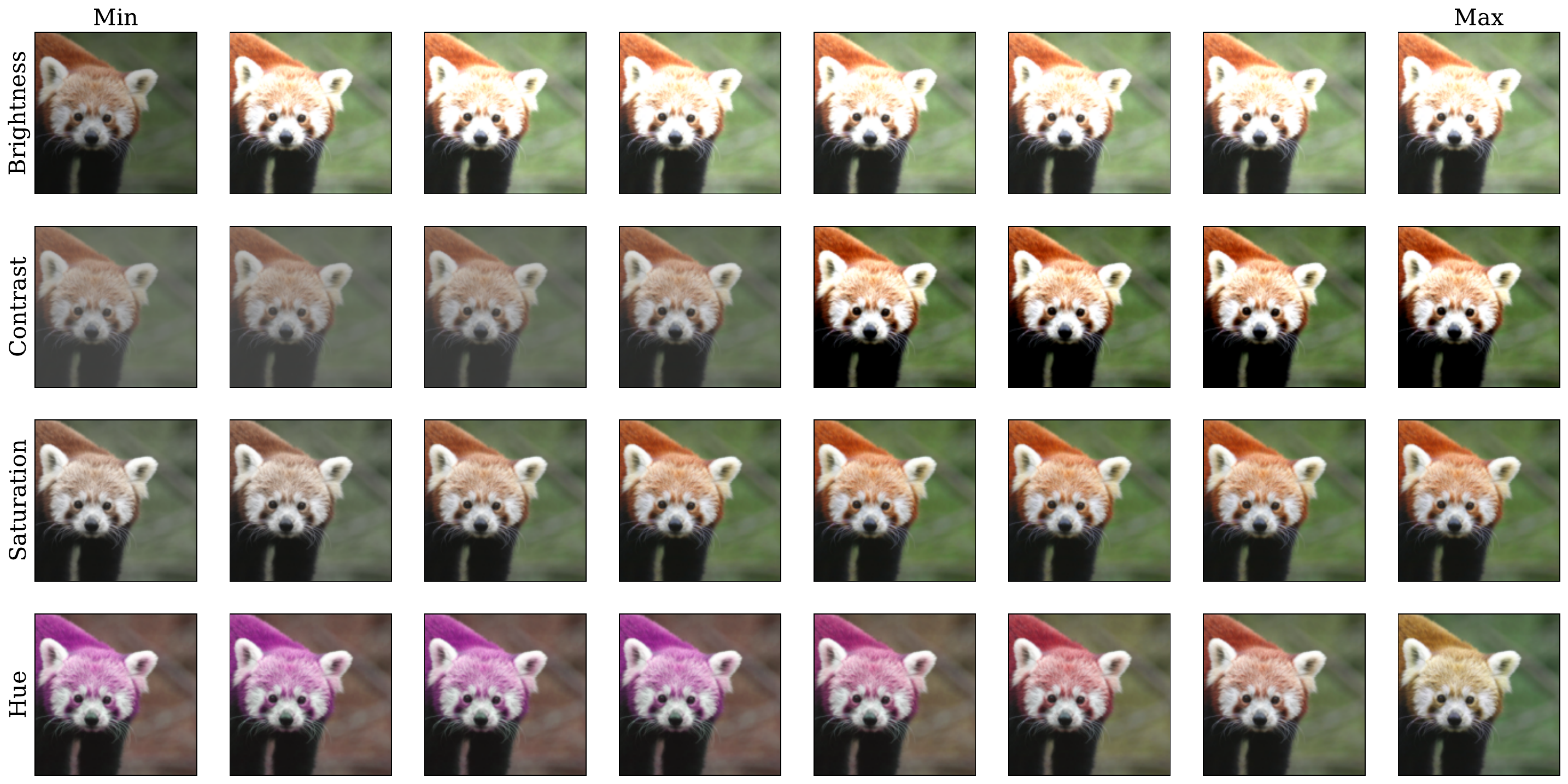}
    \caption{Application of the world model on precise transformations. For each parameter, we vary its value on a grid to see whether the model is able to predict small changes. The model is able to show the gradient of transformations, highlighting again the capabilities of the world model. We can still notice some imperfections however, as the model was only trained on combinations of augmentations. To make changes more visible, we used a model trained with a strong color jitter for this figure.}
    \label{fig:enter-label}
\end{figure}
\begin{figure}[!h]
    \centering
    \includegraphics[width=\textwidth]{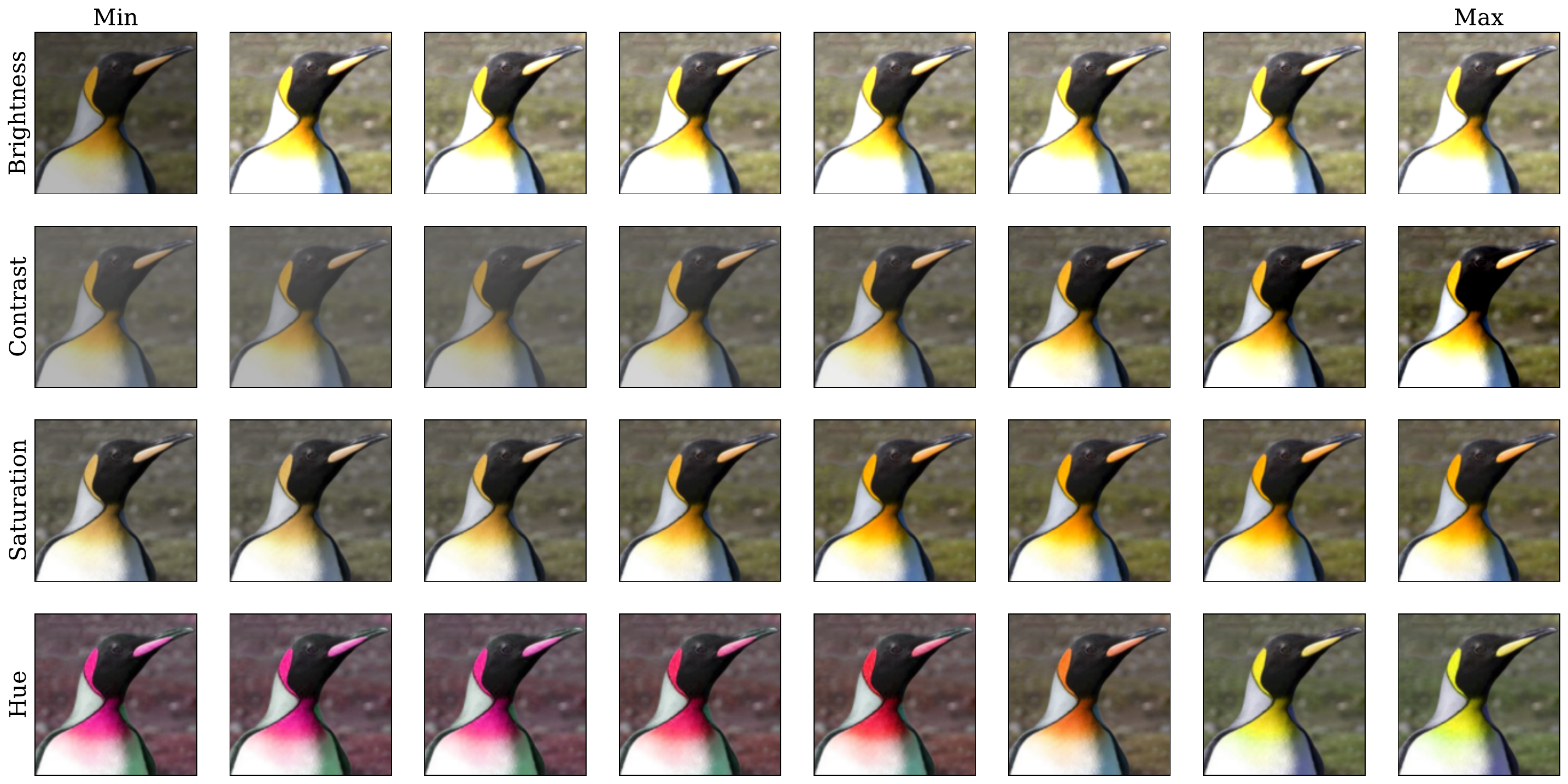}
    \caption{Application of the world model on precise transformations. For each parameter, we vary its value on a grid to see whether the model is able to predict small changes. The model is able to show the gradient of transformations, highlighting again the capabilities of the world model. We can still notice some imperfections however, as the model was only trained on combinations of augmentations. To make changes more visible, we used a model trained with a strong color jitter for this figure.}
    \label{fig:enter-label}
\end{figure}

\end{document}